\documentclass[letterpaper]{article} 
\usepackage[preprint]{aaai2027}
\usepackage[hyphens]{url}  
\usepackage{graphicx} 
\usepackage{natbib}  
\usepackage{caption} 
\usepackage{algorithm}
\usepackage{algorithmic}
\usepackage{amsmath}

\usepackage{booktabs} 
\usepackage{multirow} 
\usepackage{placeins} 
\usepackage{newfloat}
\usepackage{listings}
\DeclareCaptionStyle{ruled}{labelfont=normalfont,labelsep=colon,strut=off} 
\floatstyle{ruled}
\newfloat{listing}{tb}{lst}{}
\floatname{listing}{Listing}

\title{Intent Speaks Louder: Controllable User Simulation Beyond Response Imitation}

\author{
    Bo Wang,
    Ruixing Zhang,
    Yunqi Liu,
    Yang Zhang,\\
    Liangzhe Han,
    Tongyu Zhu,
    Leilei Sun
}
\affiliations{
    the State Key Laboratory of Complex and Critical Software Environment, Beihang University\\
    ptwang@buaa.edu.cn, yyxzhj@buaa.edu.cn,
    liuyunqi@buaa.edu.cn, yangzhang-cn@buaa.edu.cn,\\
    liangzhehan@buaa.edu.cn, tongyuzhu@buaa.edu.cn,
    leileisun@buaa.edu.cn
}

\begin{document}

\maketitle

\newcommand{\method}{UserIDA}
\newcommand{\methodfull}{User Intent-Directive Alignment}
\newcommand{\icr}{Intent-Calibrated Relative Reward}

\begin{abstract}
    User simulators are widely used as scalable environments for training
    and evaluating interactive assistants.
    Generating the next user turn is inherently one-to-many: 
    the same profile and dialogue context may support multiple plausible continuations with different local interaction intents.
    A fluent response may therefore advance the dialogue through an inappropriate
    intent, such as acceptance rather than repair.
    Our key insight is that controllable user simulation should separate
    \emph{which local interaction intent the next user turn should realize} from
    \emph{how that intent is expressed in language}.
    We introduce \method{} (\methodfull), which exposes interaction intent as an
    explicit per-turn directive.
    \method{} defines a six-way intent interface, learns directive-conditioned
    generation through supervised fine-tuning, and uses 
    intent-calibrated policy optimization during group-based reinforcement learning.
    The reward preserves composite response quality while ensuring that intent-violating
    candidates rank below compliant alternatives in mixed groups.
    On LMSYS-USP, \method{} achieves 86.6\% intent accuracy, outperforming the
    strongest dedicated user-simulator baseline by 24.3 percentage points while
    improving semantic and stylistic similarity.
    In within-context interventions, 
    it realizes at least four of the six target intents in 91.7\%
    of evaluated dialogue states, compared with 22.9\% for the strongest external baseline.
    These results establish per-turn intent control as a complementary dimension
    to response fidelity in user simulation.
\end{abstract}

\begin{links}
  \link{Code}{https://github.com/ptwang773/UserIDA}
\end{links}

\section{Introduction}

Large language models (LLMs) are increasingly deployed as interactive assistants rather than single-turn answer engines.
In real use, users reveal and revise their requests across turns: they supply
missing details, correct misunderstandings, refine constraints, and signal when a task is complete.
Evaluating and improving assistants under these conditions requires interactive environments that reproduce such behavior at scale and provide repeatable feedback for post-training.
Because real human interaction data and evaluations are costly, limited in coverage, and difficult to reproduce, LLM-based user simulators have become important infrastructure for assistant training, synthetic interaction generation, and dynamic multi-turn evaluation~\citep{balog2025usersimulation,ni2026survey,qian2025userrl,prabhakar2025apigenmt,chang2025chatbench,dou2025simulatorarena,zhou2026sim2real}.


\begin{figure}[t]
    \centering
    \includegraphics[width=\columnwidth,height=5cm, keepaspectratio]{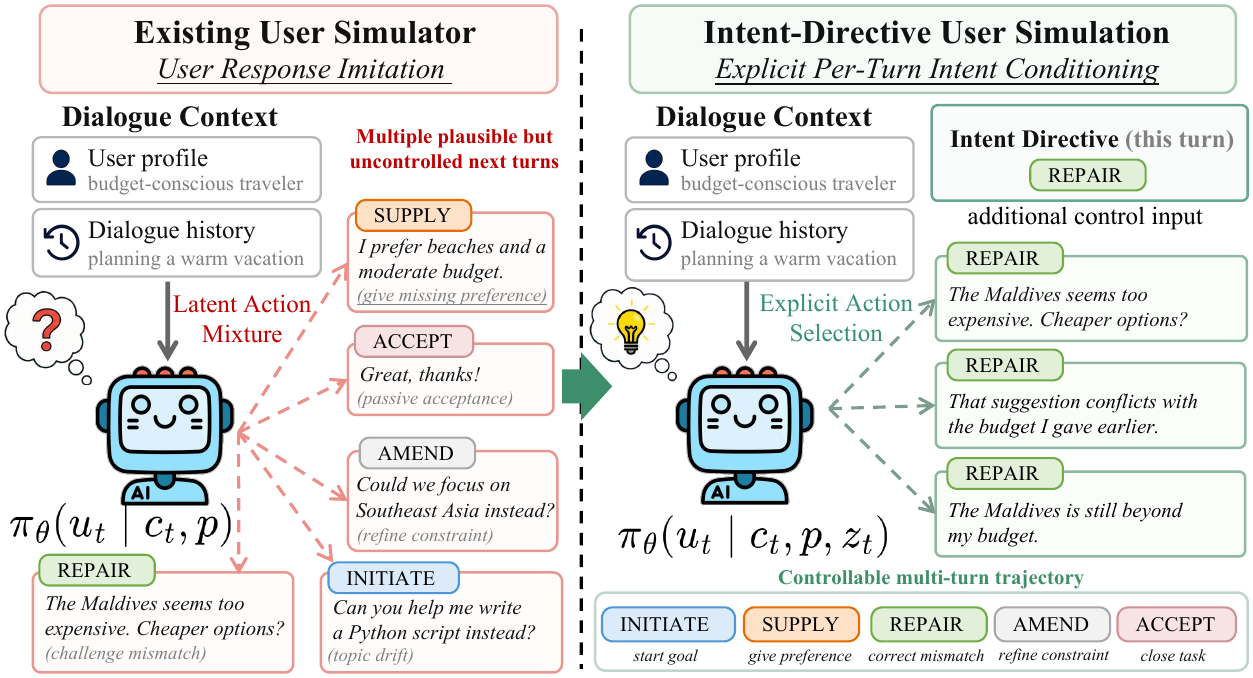}
    \caption{From response imitation to intent-directive user simulation.}
    \label{fig:intent_directive_overview}
\end{figure}

LLM-based user simulation has progressed from prompting general assistants to
role-play users~\citep{kong2024platolm,sekulic2024reliable}, supervised
adaptation of profile-conditioned or dedicated user models~\citep{
wang2025usp,naous2026flipping,wu2026humanlm}, and more recently 
reinforcement-learning objectives for goal adherence, strategic behavior, and
human-likeness~\citep{mehri2025goalalignment,zhang2026userlmr1,
wang2026turingrl}.
Across these training paradigms, however, the next local user action is usually
inferred jointly with its surface realization rather than exposed as an
explicit per-turn control variable.
A simulator may therefore produce a realistic-looking continuation that moves
the interaction in the wrong direction.
For example, after an assistant requests missing information, the simulated
user may switch topics. Similarly, after an answer violates a stated constraint, it may
accept rather than repair the mismatch.
As the left panels of Figure~\ref{fig:intent_directive_overview} illustrate, these errors change
the dialogue state and can mask assistant weaknesses in clarification, error
recovery, and task refinement.


We argue that the missing abstraction is a per-turn \emph{interaction intent},
which specifies what the next user turn should do in the conversation.
In instruction tuning, each assistant response is grounded by an explicit user instruction that serves as a local control signal.
Standard user imitation instead predicts the next user turn from dialogue history, optionally with a profile or global goal, but without an analogous per-turn directive.
The same context may support several valid continuations, including supplying
information, repairing an error, amending a request, or accepting a response,
but the simulator receives no explicit signal selecting among them.
This ambiguity motivates a turn-local control abstraction: profiles and
conversation-level goals leave the next action underspecified, whereas
target-turn paraphrases overconstrain its linguistic realization and risk
exposing content from the reference user turn.
We therefore define \emph{intent directives} as mid-level specifications of
local user actions, leaving their content, wording, and style to the simulator.

We propose \method{} (\methodfull), a framework that aligns user simulators with per-turn directives beyond response imitation.
\method{} instantiates the directive space with a canonical taxonomy covering \textsc{Initiate}, \textsc{Amend}, \textsc{Supply}, \textsc{Repair}, \textsc{SetRegister}, and \textsc{GroundAccept}, which describe how a user creates, updates, repairs, regulates, or closes the local interaction state.
Intent-SFT trains the simulator to generate the observed user turn from dialogue context, optional implicit user information, and the target directive, enabling different valid actions to be realized within the same conversation thread.
We then use intent-calibrated policy optimization to address a
quality--intent mismatch: a semantically attractive but wrong-intent candidate
may otherwise receive a favorable relative advantage.
In mixed groups, the calibration places every violating candidate below all
compliant alternatives while retaining quality discrimination among compliant
generations.
The resulting simulator can realize externally specified directives or
directive sequences for reproducible counterfactual and multi-turn evaluation.


We evaluate \method{} on real human--assistant conversations through
turn-level generation, controlled multi-turn evaluation, and within-context
directive interventions.
\method{} reaches 86.6\% turn-level intent accuracy, exceeds the strongest
dedicated simulator by 24.3 percentage points, and improves all-turn intent
success from 13\% to 58\% in controlled-prefix trajectories while preserving
semantic and stylistic fidelity.
Our contributions are as follows:
\begin{itemize}
    \item We identify interaction intent as a missing per-turn control variable
    in next-user simulation and formalize directive-conditioned generation
    through a six-way, surface-underspecified intent interface.
    \item We propose \method{}, combining directive-conditioned SFT with intent-calibrated
    policy optimization that enforces compliant-over-violating ordering
    without collapsing quality discrimination among compliant candidates.
    \item We develop turn-level, controlled multi-turn, and within-context
    evaluation protocols showing stronger intent adherence and compositional
    controllability than prompted and trained user simulators.
\end{itemize}

\section{Related Work}

\paragraph{From prompted role-play to trained user simulators.}
Classical user simulation uses agendas, goals, or dialogue acts to generate
user behavior for task-oriented dialogue policies~\citep{young2013pomdp}.
With LLMs, the field has progressed from prompting general assistants to
role-play users~\citep{kong2024platolm,sekulic2024reliable}, to supervised
training of profile-conditioned or dedicated user models~\citep{
wang2025usp,naous2026flipping,wu2026humanlm}, and to reinforcement learning for
goal adherence, strategic behavior, or human-likeness~\citep{
mehri2025goalalignment,zhang2026userlmr1,wang2026turingrl}.
These methods enrich user identity and long-horizon behavior. 
In contrast, our work focuses on a missing dimension of local execution: ensuring that each generated
user turn reliably performs the correct conversational action.

\begin{figure*}[t]
    \centering
    \includegraphics[width=0.90\textwidth]{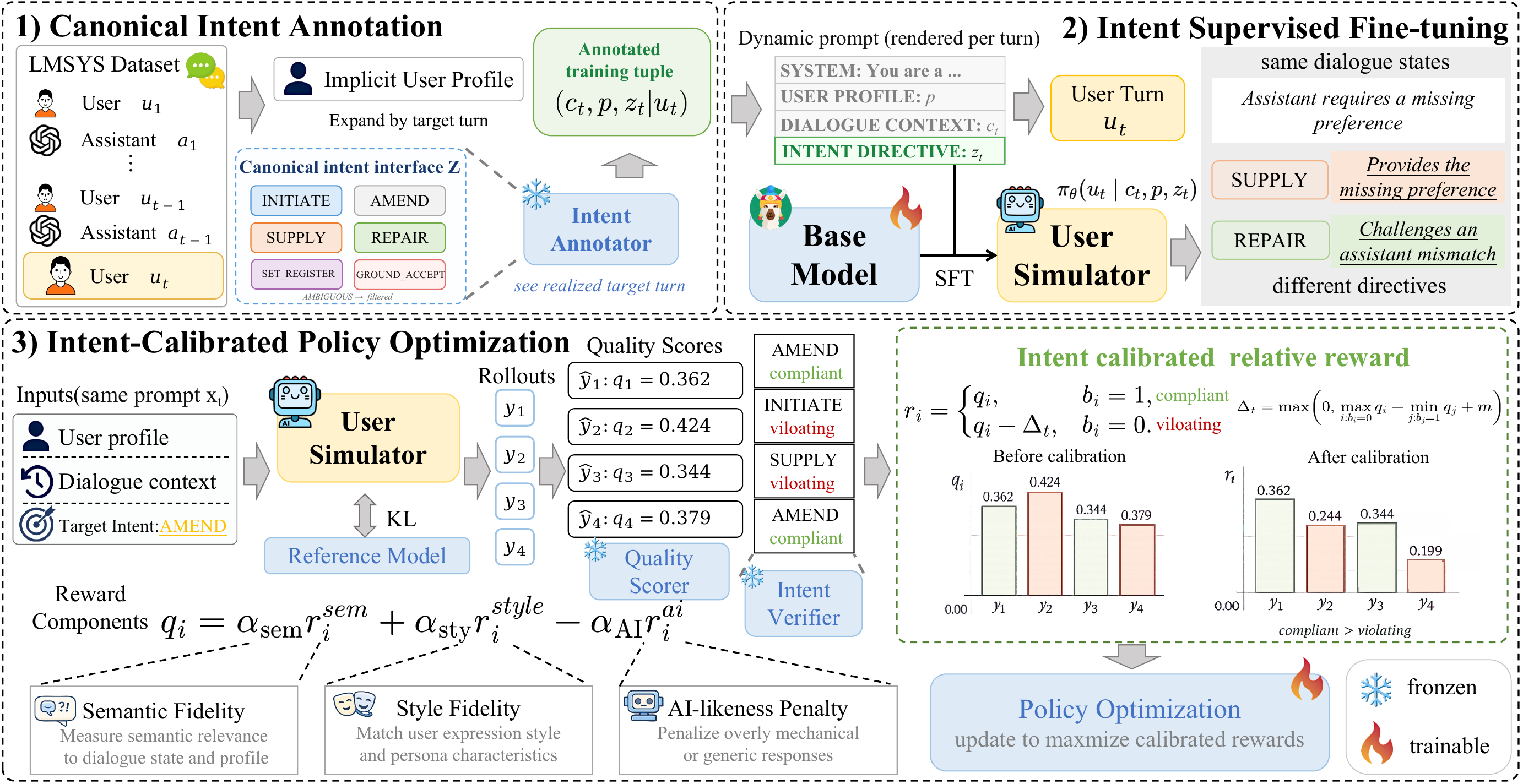}
    \caption{Overview of our proposed \methodfull{} (\method{}) framework.}
    \label{fig:method_overview}
\end{figure*}

\paragraph{Learning Objectives and Simulator Evaluation.}
Most trained simulators maximize next-turn likelihood or reference similarity. 
Recent alternatives optimize profile consistency, latent-state alignment, goal alignment, or response indistinguishability~\citep{wang2025usp,mehri2025goalalignment,wu2026humanlm,wang2026turingrl}. 
At evaluation time, automatic similarity metrics and LLM judges are commonly complemented by human evaluation, 
while SimulatorArena, ChatBench, MirrorBench, 
and recent Sim2Real studies investigate whether simulated users reproduce human interaction patterns or yield reliable assistant comparisons~\citep{chang2025chatbench,dou2025simulatorarena,hathidara2026mirrorbench,zhou2026sim2real}. 
These studies show that surface quality and general model capability do not by themselves guarantee faithful user behavior. 
Our work is complementary: we treat the requested per-turn intent as an explicit constraint, 
adapt mixed-reward calibration~\citep{liao2025rlmr} to maintain the desired groupwise ordering during RL, 
and evaluate both intent realization and complete-turn quality.


\section{Problem Formulation}
\label{sec:formulation}

\paragraph{Dialogue state and user information.}
Let a conversation be
$d=\{(u_1,a_1),\ldots,(u_T,a_T)\}$, where $u_t$ and $a_t$ are the
user and assistant turns at step $t$.
Before generating $u_t$, the visible dialogue state is
$c_t=\{(u_1,a_1),\ldots,(u_{t-1},a_{t-1})\}$, ending with the latest
assistant response.
We use $p$ for optional implicit user information.

\paragraph{Controlled intent realization.}
We define an interaction intent $z_t\in\mathcal{Z}$ as the categorical local
state transition realized by $u_t$.
A complete user policy may be factorized as
\begin{equation}
\pi(u_t\mid c_t,p)
=
\sum_{z\in\mathcal{Z}}
P(z\mid c_t,p)\,
\pi(u_t\mid c_t,p,z).
\label{eq:intent_marginalization}
\end{equation}
Standard user imitation models the marginal distribution and therefore does
not expose control over which plausible transition is generated.
We study the conditional realizer
$\pi_\theta(u_t\mid c_t,p,z_t)$, where $z_t$ specifies the intended
interaction category and the model determines context-appropriate content,
wording, and style.

\paragraph{Directive sources and scope.}
We distinguish realized-intent verification from prospective intent selection.
A frozen retrospective verifier
$g_{\mathrm{ret}}(c_t,p,v)$ assigns a canonical intent to an observed or
generated user turn $v$.
During corpus construction, $v=u_t$. During RL and automatic evaluation,
$v$ is a sampled user turn.
In controlled generation, an evaluator or scripted policy directly supplies
$z_t$.
A separate prospective predictor $h_\phi(c_t,p)$ may propose an intent without
observing $u_t$.
A directive-faithful simulator should generate a plausible user turn under
$(c_t,p)$ whose realized intent matches $z_t$.

\section{Method}
\label{sec:method}

\method{} consists of three stages: canonical intent annotation, 
intent supervised fine-tuning, 
and intent-calibrated policy optimization. 
Figure~\ref{fig:method_overview} summarizes the pipeline.

\subsection{Canonical Interaction-Intent Directives}
\label{sec:intent_taxonomy}

\begin{table}[h]
  \centering
  \small
  \setlength{\tabcolsep}{4pt}
  \begin{tabular}{p{0.29\columnwidth} p{0.62\columnwidth}}
  \toprule
  \textbf{Directive} & \textbf{Local interaction-state meaning} \\
  \midrule
  \textsc{Initiate} & Start a new top-level goal or thread. \\
  \textsc{Amend} & Refine or update an ongoing task. \\
  \textsc{Supply} & Provide requested information or a choice. \\
  \textsc{Repair} & Correct or challenge an assistant mismatch. \\
  \textsc{SetRegister} & Set the role, protocol, tone, or response mode. \\
  \textsc{GroundAccept} & Accept, acknowledge, continue, or close. \\
  \bottomrule
  \end{tabular}
  \caption{Canonical interaction-intent directives.}
  \label{tab:intent_directives_main}
\end{table}

We seek a compact, single-label interface that captures 
how a user turn changes the local state of an open-domain human--assistant dialogue. 
The labels must be turn-local, operationally distinguishable, and sufficiently abstract that they do not specify the target payload or wording. 
Table~\ref{tab:intent_directives_main} summarizes the resulting six intents. 
Our six labels define a primary control interface rather than a complete
multidimensional dialogue-act ontology.
For turns that realize multiple actions, the annotation protocol selects the
primary local state transition according to the boundary and priority rules
in Appendix~\ref{app:intent_taxonomy}.
Each annotated label conditions Intent-SFT and serves as the target constraint
during group-relative optimization.
We report label distribution and annotation quality analysis in Appendix~\ref{app:data_details}.

\subsection{Intent Supervised Fine-Tuning}
\label{sec:intent_sft}

The annotated data define
$\mathcal{D}_{\mathrm{SFT}}=\{(c_t,p,z_t,u_t)\}$.
A renderer $\Phi$ places the dialogue context, optional implicit user information, and target directive into a fixed chat template. The simulator is trained only on target user-turn tokens:
\begin{equation}
\mathcal{L}_{\mathrm{SFT}}(\theta)
=
-\sum_{(c,p,z,u)\in\mathcal{D}_{\mathrm{SFT}}}
\sum_{\ell=1}^{|u|}
\log \pi_\theta(u_\ell\mid u_{<\ell},\Phi(c,p,z)).
\label{eq:intent_sft}
\end{equation}
Conditioning on $z$ teaches a mapping from the same dialogue state to 
distinct classes of valid user behavior, 
rather than merely shifting an LM toward user-like language.
\subsection{Intent-Calibrated Policy Optimization}
\label{sec:icr_reward}

Intent-SFT substantially improves directive following, but sampled
continuations may still receive high response-quality scores while realizing
the wrong intent.
Under group-relative optimization, such a candidate can obtain a favorable
relative advantage.
We address this quality--intent mismatch by calibrating candidate rewards
before the policy update.

\paragraph{Composite response quality.}

For each directive-conditioned prompt
$x_t=\Phi(c_t,p,z_t)$, the current policy samples a group of $K$
candidate user turns:
$y_i\sim\pi_\theta(\cdot\mid x_t)$.
Let $u_t^\star$ denote the observed user turn associated with the training
instance.
We compute semantic similarity $r_i^{\mathrm{sem}}$, style similarity
$r_i^{\mathrm{style}}$, and an AI-likeness penalty $r_i^{\mathrm{ai}}$ using
frozen scorers, and define
\begin{equation}
q_i=
\alpha_{\mathrm{sem}}r_i^{\mathrm{sem}}
+\alpha_{\mathrm{sty}}r_i^{\mathrm{style}}
-\alpha_{\mathrm{AI}}r_i^{\mathrm{ai}}.
\label{eq:quality_reward}
\end{equation}
These fixed coefficients combine response-quality criteria only.
Intent compliance is imposed separately through the relative calibration below.

\paragraph{Why a fixed intent bonus is insufficient.}
A natural alternative is a fixed mixture
$\widetilde r_i=q_i+\lambda b_i$.
For a compliant candidate $c$ and a violating candidate $v$,
$\widetilde r_v>\widetilde r_c$ whenever
$q_v-q_c>\lambda$.
Thus, any fixed bonus can be overwhelmed by variation in the scale or spread
of the quality scorer.
Our calibration instead computes the minimum group-specific shift required to
separate the compliant and violating sets.

\paragraph{Intent-calibrated relative reward.}
For each candidate, a frozen canonical verifier predicts the realized
interaction intent:
$\hat{z}_i=g_{\mathrm{ret}}(c_t,p,y_i)$.
We define exact directive compliance as
$b_i=\mathbf{1}\!\left[\hat{z}_i=z_t\right]$.
Candidates with $b_i=1$ are intent-compliant, whereas candidates with
$b_i=0$ are intent-violating.
The informative case is a \emph{mixed group}, containing at least one
compliant and one violating candidate.
We compute the group-specific calibration magnitude
\begin{equation}
\Delta_t
=
\max\!\left(
0,\,
\max_{i:b_i=0}q_i
-
\min_{j:b_j=1}q_j
+
m
\right),
\end{equation}
where $m>0$ is a fixed margin.
The calibrated reward is
\begin{equation}
r_i=
\begin{cases}
q_i, & b_i=1,\\
q_i-\Delta_t, & b_i=0.
\end{cases}
\end{equation}
This construction guarantees
$\max_{i:b_i=0}r_i + m
\leq
\min_{j:b_j=1}r_j$
for every mixed group.
For an all-compliant group, we set $r_i=q_i$ because the intent constraint is
already satisfied.
For an all-violating group, we also use $r_i=q_i$: without a compliant
alternative, the group provides no relative intent preference, and subtracting
a uniform penalty would be removed by groupwise centering.
We record the all-violating-group rate as a diagnostic of insufficient
directive exploration.
Thus, no high-quality intent-violating candidate can outrank an
intent-compliant alternative after calibration.
Within the compliant subset, the composite quality score continues to
distinguish semantic fidelity, stylistic alignment, and user-like expression.

\paragraph{Group-relative policy update.}
Following Group Relative Policy Optimization (GRPO)~\citep{shao2024deepseekmath}, rewards are standardized within
each sampled group:
\begin{equation}
\widehat A_i
=
\frac{r_i-\bar r}
{\sqrt{s_r^2+\varepsilon_{\mathrm{num}}}},
\quad
\bar r=\frac{1}{K}\sum_{j=1}^{K}r_j.
\label{eq:grpo_group_advantage_main}
\end{equation}
For token $\ell$ of $y_i$, let
$\rho_{i,\ell}(\theta)=
\pi_\theta(y_{i,\ell}\mid x_t,y_{i,<\ell})/
\pi_{\theta_{\mathrm{old}}}(y_{i,\ell}\mid x_t,y_{i,<\ell})$ and
$
\mathcal{L}^{\mathrm{clip}}_{i,\ell}
=
\min\!\left(
\rho_{i,\ell}\widehat A_i,\,
\operatorname{clip}(\rho_{i,\ell},1-\epsilon,1+\epsilon)\widehat A_i
\right)$.
We maximize
\begin{equation}
\mathcal{J}(\theta)
=
\mathbf{E}\!\left[
\frac{1}{K}\sum_{i=1}^{K}\frac{1}{|y_i|}
\sum_{\ell=1}^{|y_i|}
\left(
\mathcal{L}^{\mathrm{clip}}_{i,\ell}
-\beta d^{\mathrm{KL}}_{i,\ell}
\right)
\right].
\label{eq:grpo_objective_main}
\end{equation}
Here $d^{\mathrm{KL}}_{i,\ell}$ is the per-token divergence from the frozen
reference policy and $\beta$ controls its strength.

Group standardization is strictly increasing whenever the reward variance is
nonzero, so our Intent-calibrated Policy Optimization implies that every violating
candidate has a lower normalized advantage than every compliant alternative.
This is an ordering guarantee, not a guarantee that every violating advantage
is negative.
Appendix~\ref{app:grpo_details} gives the full derivation and implementation
details.

\section{Experiments}
\label{sec:experiments}

\subsection{Experimental Setup}
\label{sec:exp_setup}

\paragraph{Training Data.}
We use LMSYS-Chat-1M~\citep{zheng2024lmsys}, a large-scale collection of
human--LLM conversations.
Following the preprocessing protocol of USP~\citep{wang2025usp}, we remove
non-English, toxic, redundant, and low-quality conversations.
Expanding into next-user-turn instances yields 444,635 non-ambiguous training turns, 
27,778 validation turns, and 9,233 test turns.
We retain the USP-style implicit user profiles associated with these
conversations, which summarize persistent user attributes, preferences, and
interaction tendencies without directly exposing future user utterances (Appendix~\ref{app:data_details}).


\begin{table*}[t]
    \centering
    \small
    \setlength{\tabcolsep}{5.5pt}
    \begin{tabular}{@{}llcccccc@{}}
    \toprule
    \textbf{Family} &
    \textbf{Model / Interface} &
    \textbf{Intent Acc.}$\uparrow$ &
    \textbf{Macro-F1}$\uparrow$ &
    \textbf{SimCSE}$\uparrow$ &
    \textbf{StyleCSE}$\uparrow$ &
    \textbf{Ctx. Valid.}$\uparrow$ &
    \textbf{User Auth.}$\uparrow$ \\
    \midrule

    \multirow{6}{*}{Base LMs}
    & LLaMA-3-8B Base, w/o Directive
    & 40.30 & 0.219 & 0.290 & 0.278 & 24.90 & 23.45 \\
    & LLaMA-3-8B Base, w/ Directive
    & 56.20 & 0.541 & 0.285 & 0.265 & 28.51 & 26.03 \\
    & Gemini-2.5-Flash, w/o Directive
    & 50.65 & 0.333 & 0.492 & 0.170 & 80.95 & 89.67 \\
    & Gemini-2.5-Flash, w/ Directive
    & 75.97 & 0.746 & 0.512 & 0.177 & 83.16 & 89.29 \\
    & GPT-4o, w/o Directive
    & 51.18 & 0.315 & 0.455 & 0.141 & 80.83 & 89.67 \\
    & GPT-4o, w/ Directive
    & 77.34 & 0.721 & 0.475 & 0.146
    & \underline{86.16} & \underline{89.73} \\
    \midrule

    \multirow{4}{*}{User simulators}
    & UserLM, w/o Directive
    & 36.65 & 0.258 & 0.306 & 0.248 & 44.85 & 50.05 \\
    & UserLM, w/ Directive
    & 44.94 & 0.345 & 0.309 & 0.232 & 44.79 & 48.12 \\
    & USP, w/o Directive
    & 56.27 & 0.406 & 0.475 & 0.397 & 72.57 & 79.24 \\
    & USP, w/ Directive
    & 62.28 & 0.509 & 0.494 & 0.407 & 73.77 & 79.92 \\
    \midrule

    \multirow{2}{*}{Ours}
    & \method{} w/o RL
    & \underline{81.98} & \underline{0.822}
    & \underline{0.574} & \underline{0.469}
    & 81.56 & 83.98 \\
    & \method{}
    & \textbf{86.62} & \textbf{0.864}
    & \textbf{0.591} & \textbf{0.476}
    & \textbf{86.73} & \textbf{89.89} \\
    \bottomrule
    \end{tabular}
    \caption{Turn-level next-user simulation.}
    \label{tab:main_turn_level}
\end{table*}

\paragraph{Intent Directive Annotation.}
We use one frozen Qwen3.5-9B retrospective verifier
$g_{\mathrm{ret}}$~\citep{qwen2026qwen35} for corpus annotation, rollout
compliance, and automatic intent evaluation.
Given the implicit profile, visible dialogue state, and an observed or
generated user turn, it assigns exactly one of the six canonical directives.
For corpus construction, the observed target turn is available only to
$g_{\mathrm{ret}}$ and serves as the supervised output.
All systems in the controlled experiments receive the same target directive.
We separately evaluate a prospective Qwen3.5-9B predictor
$h_\phi(c_t,p)$ that does not observe the next user turn.
This experiment assesses whether the directive interface can be driven autonomously.
Agreement with expert labels, including evaluation on generated user turns,
is reported in Appendix~\ref{app:data_details}.


\paragraph{Baselines.}
We compare against general LMs (LLaMA-3-8B Base, GPT-4o, and
Gemini-2.5-Flash)~\citep{grattafiori2024llama3,openai2024gpt4o,
comanici2025gemini25}
and dedicated simulators USP-8B~\citep{wang2025usp} and UserLM-8B~\citep{naous2026flipping}. 
Each prompted model is evaluated with matched six-shot interfaces with and without the target directive. 
Profile, context, demonstrations, decoding, and output format are otherwise identical. 
\method{} without RL denotes the Intent-SFT checkpoint.


\paragraph{Metrics.}
We report three groups of turn-level metrics.
\textbf{Intent Accuracy} measures exact agreement between the requested and realized intent; 
\textbf{Macro-F1}, reported on a 0--1 scale, reduces the effect of class imbalance.
\textbf{SimCSE}~\citep{gao2021simcse} and \textbf{StyleCSE} measure semantic 
and stylistic similarity to the observed user turn. 
A frozen, reference-free LLM judge that does not observe the gold user turn
scores \textbf{Contextual Validity} and \textbf{User Authenticity}.
Contextual Validity combines intent realization, dialogue-state coherence,
and information appropriateness, whereas User Authenticity combines profile
consistency with natural and economical user-side expression.
Full metric definitions and evaluation prompts are provided in
Appendices~\ref{app:additional_results} and~\ref{app:prompts}.

\paragraph{Implementation.}
We initialize all trained variants from LLaMA-3-8B Base. 
Intent-SFT trains for three epochs with LoRA ($r=64$, $\alpha=32$)~\citep{hu2022lora}, 
learning rate $5\times10^{-5}$, and maximum length 4,096. 
Group-relative optimization uses $K=4$, margin $m=0.10$, LoRA ($r=16$, $\alpha=32$), 
learning rate $5\times10^{-7}$, and KL coefficient $0.05$. 
Training uses BF16 on four RTX 4090 GPUs. 
We optimize both stages with AdamW~\citep{loshchilov2019decoupled}.
Appendix~\ref{app:implementation} provides complete optimization, 
decoding, software, and compute details.
The RL quality score combines semantic similarity, style similarity, and an
AI-likeness penalty with fixed weights $(0.50,0.02,0.05)$, respectively.


\subsection{Turn-Level Main Results}
\label{sec:main_results}
\begin{figure}[t]
    \centering
    \includegraphics[width=0.95\columnwidth]
    {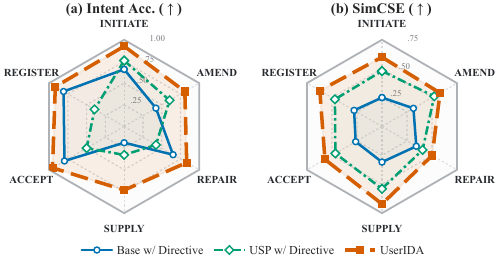}
    \caption{Per-intent turn-level results.}
    \label{fig:intent_radar}
\end{figure}

We report the main turn-level results in Table~\ref{tab:main_turn_level} and Figure~\ref{fig:intent_radar}, with the following conclusions:
\paragraph{Intent directives help, prompting alone is insufficient.}
Adding the directive improves intent accuracy by 6.01--26.16 percentage points
across matched prompted baselines.
Nevertheless, the strongest prompted dedicated simulator, USP, reaches only
62.28\%, compared with 81.98\% for Intent-SFT.
This gap shows that exposing the interface at inference time is useful, but
learning directive-conditioned realization is substantially more effective.

\paragraph{\method{} provides the strongest control--quality balance.}
\method{} reaches 86.62\% intent accuracy and 0.864 macro-F1, exceeding the
strongest general-model baseline by 9.28 points and USP by 24.34 points.
It obtains the best point estimates for SimCSE, StyleCSE, Contextual
Validity, and User Authenticity.
USP's strong style score confirms that profile fidelity and local intent
control are complementary: profile conditioning helps preserve how a user
speaks, whereas directive training is needed to control what the turn does.

\paragraph{The gains extend to minority intents.}
Although \textsc{Initiate} and \textsc{Amend} account for 81.4\% of labeled
turns, Figure~\ref{fig:intent_radar} shows that \method{} achieves the highest
accuracy on all six directives.
For \textsc{Supply}, which constitutes only 1.9\% of labeled turns, accuracy
increases from 32.8\% for USP with directives to 73.4\% for \method{}.
Thus, the aggregate gain is not explained only by the majority classes.

\begin{table*}[t]
    \centering
    \small
    \setlength{\tabcolsep}{3.4pt}
    \begin{tabular}{@{}lccccccccc@{}}
    \toprule
    \textbf{Model / Interface} &
    \textbf{Intent Acc.}$\uparrow$ &
    \textbf{Mean Traj. Acc.}$\uparrow$ &
    \textbf{All-Turn}$\uparrow$ &
    \textbf{SimCSE}$\uparrow$ &
    \textbf{StyleCSE}$\uparrow$ &
    \textbf{Role}$\uparrow$ &
    \textbf{Int.}$\uparrow$ &
    \textbf{Goal}$\uparrow$ &
    \textbf{Total}$\uparrow$ \\
    \midrule
    USP, w/ Directive
    & 60.17 & 59.43 & 13.00 & 0.480 & 0.386
    & 65.70 & 60.90 & 64.30 & 63.63 \\
    UserLM, w/ Directive
    & 41.70 & 40.12 & 2.00 & 0.304 & 0.272
    & \underline{75.40} & 68.80 & 71.90 & \underline{72.03} \\
    \method{} w/o RL
    & \underline{82.37} & \underline{83.00}
    & \underline{48.00} & \underline{0.562} & \underline{0.439}
    & 73.00 & \underline{70.20} & \underline{72.70} & 71.97 \\
    \method{}
    & \textbf{88.38} & \textbf{88.23}
    & \textbf{58.00} & \textbf{0.584} & \textbf{0.450}
    & \textbf{80.10} & \textbf{76.20} & \textbf{80.50}
    & \textbf{78.93} \\
    \bottomrule
    \end{tabular}
    \caption{Conversation-level performance comparison of different user simulators.}
    \label{tab:multiturn_results}
\end{table*}

\subsection{Ablation and Optimization Analysis}
\label{sec:ablation}

We compare four settings to isolate directive conditioning, supervised
adaptation, and intent calibration.
Starting from the directive-prompted base model, we introduce Intent-SFT (\method{} w/o RL) and
then continue from the same supervised checkpoint with either quality-only
GRPO or the full \method{} objective.
Quality-only GRPO uses the same composite quality score $q_i$ and the
same group-based policy objective as \method{}, but removes intent
calibration by setting $r_i=q_i$ for every candidate.
This comparison isolates \icr{} from the semantic, stylistic, and AI-likeness
quality components.

\begin{figure}[t]
    \centering
    \includegraphics[width=0.8\columnwidth]
    {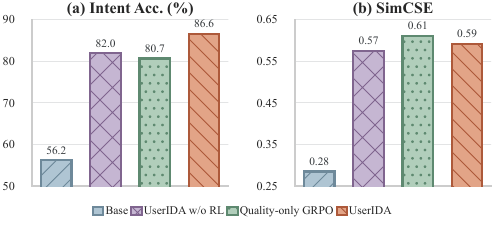}
    \caption{
    Effect of supervised directive conditioning and intent calibration.
    }
    \label{fig:reward_ablation}
\end{figure}

As shown in Figure~\ref{fig:reward_ablation}, 
adding Intent-SFT raises intent accuracy from 56.2\% to 82.0\% and SimCSE from 0.28 to 0.57.
Quality-only GRPO further raises SimCSE to 0.61 but lowers intent accuracy to 80.7\%, directly demonstrating the quality-intent mismatch motivating \icr{}. 
Full \method{} reaches 86.6\% intent accuracy and 0.59 SimCSE. 
It improves both metrics over supervised training and recovers 5.9 points of intent accuracy over quality-only RL while retaining most of its semantic gain. 


\begin{table*}[t]
    \centering
    \small
    \setlength{\tabcolsep}{5pt}
    \begin{tabular}{lcccccc}
    \toprule
    \textbf{Model / Interface}
    & \textbf{Intent Acc.}$\uparrow$
    & \textbf{Macro-F1}$\uparrow$
    & \textbf{4-of-6 Succ.}$\uparrow$
    & \textbf{Coverage}$\uparrow$
    & \textbf{Ctx. Valid.}$\uparrow$
    & \textbf{User Auth.}$\uparrow$ \\
    \midrule
    LLaMA-3-8B Base w/ Directive
    & 41.32 & 0.424 & 22.92 & 3.56 & 26.81 & 25.64 \\
    USP w/ Directive
    & 22.92 & 0.202 & 2.08 & 1.83 & 66.76 & 83.44 \\
    UserLM w/ Directive
    & 20.49 & 0.182 & 2.08 & 1.90 & 32.08 & 39.31 \\
    \method{} w/o RL
    & \underline{66.32} & \underline{0.661} & \underline{72.92}
    & \underline{4.29} & \underline{79.03} & \textbf{83.96} \\
    \method{}
    & \textbf{75.35} & \textbf{0.740} & \textbf{91.67}
    & \textbf{4.71} & \textbf{80.66} & \underline{83.91} \\
    \bottomrule
    \end{tabular}
    \caption{Within-context directive intervention.}
    \label{tab:counterfactual_control}
\end{table*}

\subsection{Controlled Multi-Turn Trajectory Evaluation}
\label{sec:multiturn_eval}

We evaluate whether local intent control composes across multiple checkpoints in the same multi-turn dialogue. 
At each checkpoint, every model receives the same gold profile and gold dialogue prefix ending at the preceding assistant response.
Generated turns are not recursively fed into later checkpoints. 
This controlled-prefix protocol isolates multi-turn consistency from divergence in assistant responses and generated histories.

\textbf{Step Intent Accuracy} micro-averages intent correctness over all
checkpoints.
\textbf{Mean Trajectory Accuracy} first averages within each trajectory, and
\textbf{All-Turn Success} requires every evaluated turn in a trajectory to
realize the requested intent.
Role, Interaction, and Goal are reference-free trajectory-level rubric scores.
Their full definitions are provided in Appendix~\ref{app:additional_results}.

As shown in Table~\ref{tab:multiturn_results}, compared with the strongest external simulator for intent control, 
USP w/ Directive, \method{} improves step intent accuracy by 28.21 points and all-turn success by 45 points. 
UserLM receives stronger role-based rubric scores than USP but realizes all requested intents in only 2\% of trajectories, 
illustrating that role plausibility and local control are distinct. 
\method{} leads both dimensions, reaching 58\% all-turn success and a total trajectory score of 78.93. 
The gain over \method{} w/o RL is also consistent across intent, similarity, and rubric metrics. 
These results show that trained directive control composes more reliably across
changing dialogue states than few-shot prompting alone.

\subsection{Within-Context Directive Intervention}
\label{sec:counterfactual_control}
The natural test set evaluates different intents in different dialogue states,
making it difficult to isolate the effect of the control signal itself.
We therefore construct a within-context intervention suite.
We first conduct a human affordance screen and retain only dialogue states for
which annotators judge all six canonical directives to be interactionally
feasible under the visible state.
All systems receive the same retained dialogue states and generate one
next-user turn for each of the six target directives.
In addition to Intent Accuracy and Macro-F1, 
4-of-6 Success is the percentage of contexts in which at least four requested
intents are correctly realized.
Coverage is the mean number of distinct realized intent labels per context.
Full construction and metric definitions are provided in
Appendix~\ref{app:additional_results}.

As shown in Table~\ref{tab:counterfactual_control}, \method{} achieves the
strongest within-context control.
It exceeds the strongest external baseline by 34.03 percentage points in
Intent Accuracy and increases the fraction of contexts with at least four
correct interventions from 22.92\% to 91.67\%.
Relative to \method{} w/o RL, intent accuracy increases from 66.32\%
to 75.35\%, with a paired-bootstrap improvement of 9.03 points
(95\% CI: [4.17, 14.24]).
Realized-intent coverage also rises from 4.29 to 4.71, while Contextual
Validity remains comparable and User Authenticity is unchanged.
The gain therefore reflects stronger control over the local interaction
transition rather than a reduction to rigid or unnatural templates.
\begin{figure}[t]
    \centering
    \includegraphics[width=\columnwidth]
    {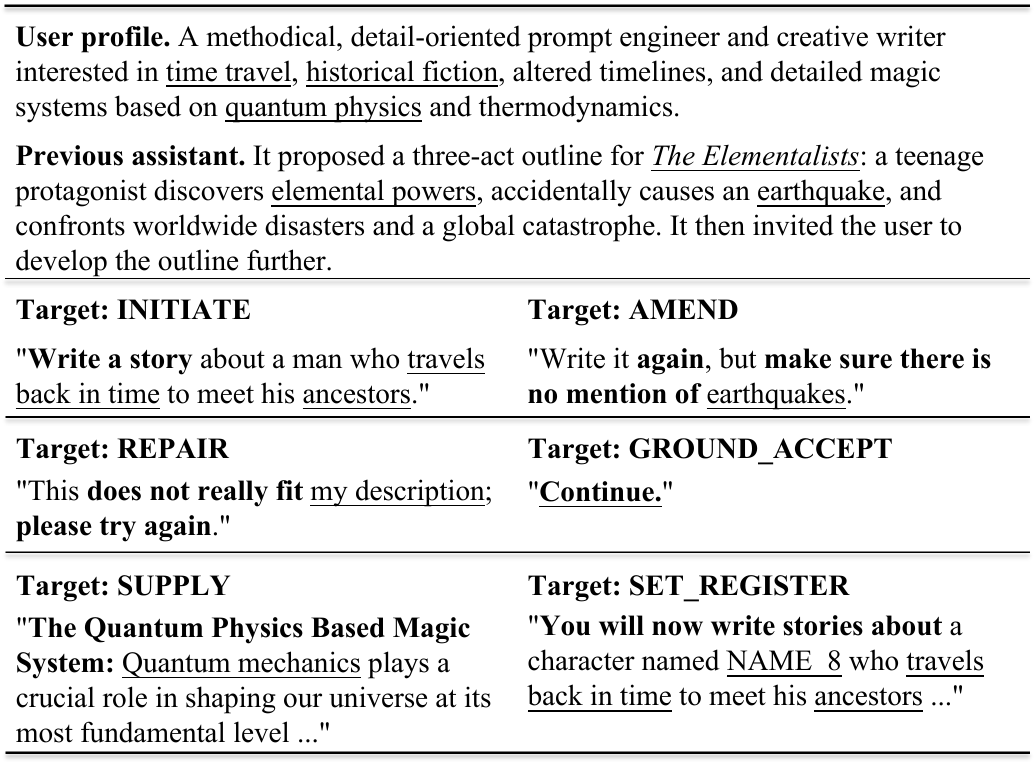}
    \caption{Within-context intervention case.
     Boldface marks spans realizing the requested local interaction, while
     underlining marks content grounded in the profile or dialogue state.
     }
    \label{fig:counterfactual_case}
\end{figure}

Figure~\ref{fig:counterfactual_case} illustrates the capability measured
quantitatively in Table~\ref{tab:counterfactual_control}.
From one fixed dialogue state, changing only the directive produces
interactionally distinct continuations: a new goal under \textsc{Initiate},
a revision under \textsc{Amend}, an explicit correction under
\textsc{Repair}, an acknowledgment under \textsc{GroundAccept}, a payload
under \textsc{Supply}, and an interaction protocol under
\textsc{SetRegister}.
This demonstrates that the directive controls the local dialogue transition
without prescribing a single surface response.
Appendix~\ref{app:additional_results} provides a matched qualitative
comparison.
Under the same dialogue state, USP w/ Directive frequently collapses multiple
directives to generic continuation requests, whereas \method{} produces
distinct, directive-consistent local transitions.

\subsection{Context-Dependence Probe}
We next test whether the six human-validated intents exhibit distinct
dependencies on the observable dialogue state.
The intent label is used only to group examples and is never provided to the
language model.
Let the full input view contain three sources:
the implicit profile $p$, the earlier dialogue context
$h_t=(u_1,a_1,\ldots,u_{t-2},a_{t-2},u_{t-1})$, and the immediately
preceding assistant turn $a_{t-1}$.
Thus, the full view is $(p,h_t,a_{t-1})$.
We evaluate three matched removals:
1) \textbf{No Profile}: $(\emptyset,h_t,a_{t-1})$;
2) \textbf{No Previous Assistant}: $(p,h_t,\emptyset)$;
3) \textbf{No Earlier History}: $(p,\emptyset,a_{t-1})$.
We also include two reduced-context controls:
\textbf{Profile Only}: $(p,\emptyset,\emptyset)$;
\textbf{User-History Only}: $(\emptyset,u_{<t},\emptyset)$,
where $\emptyset$ denotes an omitted component and
$u_{<t}=(u_1,\ldots,u_{t-1})$ retains only previous user turns.

For observed user tokens
$u_i=(u_{i,1},\ldots,u_{i,L_i})$ under input view $v$, we compute
$\mathrm{NLL}_{i,v}
=
-\frac{1}{L_i}
\sum_{\ell=1}^{L_i}
\log
\pi_0
\left(
u_{i,\ell}
\mid
u_{i,<\ell},
\Phi_v(c_i,p_i)
\right)$.
We report
$\Delta\mathrm{NLL}_{i,v}
=
\mathrm{NLL}_{i,v}
-
\mathrm{NLL}_{i,\mathrm{full}}$.
A positive value means that the corresponding removal makes the observed user
turn harder to predict.

\begin{figure}[h]
    \centering
    \includegraphics[width=0.94\columnwidth]
    {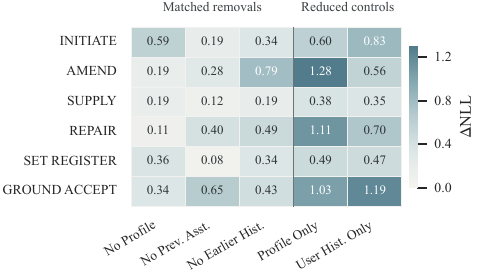}
    \caption{Directive-specific context dependence.}
    \label{fig:nll_context_heatmap}
\end{figure}
Figure~\ref{fig:nll_context_heatmap} reveals substantial heterogeneity across
directives.
\textsc{Amend} is most sensitive to earlier history
($\Delta\mathrm{NLL}=0.786$), consistent with an open task established across
turns.
\textsc{Repair} depends on both the preceding assistant response ($0.403$) and
earlier history ($0.494$), while \textsc{GroundAccept} is especially sensitive
to the preceding assistant response ($0.651$).
\textsc{Initiate} and \textsc{SetRegister} show stronger profile dependence
($0.590$ and $0.361$).
Together with the human-alignment results, these contextual signatures support
the six intents as operational interaction-state abstractions rather than
surface-only categories.

\subsection{Human Alignment and Prospective Predictability}
We compare two Qwen3.5-9B intent modes against expert reference labels.
The \emph{retrospective verifier} observes the realized user turn and
classifies its intent, whereas the \emph{prospective predictor} receives only
the profile and visible dialogue state and predicts the next observed intent
before the user turn is available.
\begin{table}[h]
\centering
\small
\setlength{\tabcolsep}{4.2pt}
\begin{tabular}{@{}lcccc@{}}
\toprule
\textbf{Qwen mode}
& \textbf{Inputs}
& \textbf{Acc.}
& \textbf{Macro-F1}
& \textbf{$\kappa$} \\
\midrule
Retrospective verifier
& $p,c_t,u_t$
& \textbf{87.88}
& \textbf{0.875}
& \textbf{0.8223} \\
Prospective predictor
& $p,c_t$
& 71.39
& 0.718
& 0.5950 \\
\bottomrule
\end{tabular}
\caption{Agreement with expert intent labels.}
\label{tab:intent_annotation_prediction}
\end{table}
The retrospective verifier reaches 87.88\% accuracy, 0.875 macro-F1, and
verifier--human agreement of $\kappa=0.8223$.
Its per-intent F1 ranges from 0.786 for \textsc{Supply} to 0.920 for
\textsc{GroundAccept}, indicating strong but non-perfect agreement with expert
judgments.
Without access to $u_t$, the prospective predictor still reaches 71.39\%
accuracy and 0.718 macro-F1.
This result shows that the observed next intent is partially recoverable from
the visible dialogue state, while the remaining gap is consistent with the
one-to-many nature of next-user behavior.
Detailed per-intent agreement and generated-output results are reported in
Appendix~\ref{app:data_details}.

\section{Conclusion}
\label{sec:conclusion}
We introduced \method{}, a controllable user-simulation framework that
separates the intended local interaction transition from its linguistic
realization.
\method{} combines a six-way intent interface, Intent-SFT, and
intent-calibrated group-relative optimization.
Across turn-level generation, controlled multi-turn evaluation, and
within-context intervention, \method{} improves intent adherence and
compositional controllability while preserving semantic, stylistic, and
reference-free interaction quality.
The quality-only GRPO ablation further shows that response similarity can
improve while intent control degrades, motivating explicit calibration between
compliant and violating candidates.
These findings establish local interaction intent as a complementary control
dimension beyond response imitation.

\bibliography{aaai2027}

\onecolumn
\appendix
\setcounter{secnumdepth}{2}

\section{Canonical Intent Directive Taxonomy and Annotation Protocol}
\label{app:intent_taxonomy}
\subsection{Design Principles}

We adopt four design principles. First, directives are \emph{turn-local}: they describe
what the next user turn does to the current interaction state rather than the user's
full persona or long-term objective. Second, they are \emph{surface-underspecified}: a directive
must not paraphrase the target turn or reveal its specific payload. Third, they are
\emph{operational}: each label has explicit decision boundaries that can be applied by
both human annotators and a frozen verifier. Fourth, they are \emph{open-domain}: the
labels characterize interaction dynamics rather than domain-specific tasks.


\subsection{Full Directive Definitions}

\begin{table}[H]
\centering
\small
\setlength{\tabcolsep}{4pt}
\begin{tabular}{p{0.115\textwidth} p{0.195\textwidth} p{0.415\textwidth} p{0.195\textwidth}}
\toprule
\textbf{Directive}
& \textbf{Interaction-state meaning}
& \textbf{Definition and boundary rule}
& \textbf{Example} \\
\midrule

\textsc{Initiate}
& Create a new top-level goal thread.
& The user opens a new task, question, request, capability probe, or open-ended discussion that is not a continuation of the current open task. It is used when none of the more specific directives apply.
& ``Can you help me plan a warm vacation?'' \\

\textsc{Amend}
& Update the state of an ongoing task.
& The user advances, refines, extends, or changes the constraints, scope, format, style, length, focus, or next step of an existing task, without explicitly claiming that the assistant was wrong. A next item, next example, next trial, or next procedural step within an open session is also \textsc{Amend}.
& ``Could we focus on Southeast Asia instead?'' \\

\textsc{Supply}
& Inject requested evidence, material, or missing information into the current dialogue state.
& The user mainly provides information, evidence, code, logs, text, numbers, preferences, choices, or answers, without issuing a new processing request. If the user provides material and simultaneously requests a new operation, the requested task action takes priority.
& ``I prefer beaches with a moderate budget.'' \\

\textsc{Repair}
& Correct or challenge an assistant-induced error state.
& The user explicitly indicates that the previous assistant response, code, solution, interpretation, or behavior was wrong, failed, misunderstood the request, or violated a requirement. A bare traceback or error log is not automatically \textsc{Repair} unless it is explicitly tied to the assistant's prior output.
& ``That option is too expensive for the budget I mentioned.'' \\

\textsc{SetRegister}
& Modify the interaction protocol, role, style, or future response register.
& The user sets the assistant's role, identity, protocol, tone, behavior mode, or response register for the current or future interaction. A purely local format constraint is not \textsc{SetRegister} unless it frames how the assistant should behave across the interaction.
& ``Act as a strict writing coach and be direct.'' \\

\textsc{GroundAccept}
& Ground, accept, acknowledge, continue, or close the current state without adding a new constraint.
& The user greets, thanks, accepts, confirms, closes, provides simple positive feedback, or asks to continue without introducing a new task constraint. If positive feedback is followed by a modification, the turn is \textsc{Amend}.
& ``Great, thanks!'' \\

\textsc{Ambiguous}
& Annotation fallback only.
& Used only when the turn cannot be stably assigned after all priority and boundary rules have been applied. \textsc{Ambiguous} turns are excluded from the six-way intent-control training set.
& ``Okay, maybe,'' when the available context does not support a stable intent assignment. \\

\bottomrule
\end{tabular}
\caption{Full canonical interaction-intent directive taxonomy.
The main training and evaluation interface uses the six canonical directives.
\textsc{Ambiguous} is retained only as an annotation fallback.
}
\label{tab:intent_directives_full}
\end{table}

\subsection{Canonical Annotation Prompt}
\label{app:intent_annotation_prompt}
The classifier identifies the primary local interaction action performed by the target
user turn. It uses the profile and dialogue prefix only to reconstruct the active task
state and the target turn's relation to the preceding assistant response. 

\begin{lstlisting}[
    caption={Canonical interaction-intent classifier prompt.},
    label={lst:intent_labeler_prompt},
    basicstyle=\ttfamily\small,
    breaklines=true,
    breakindent=0pt,
    breakatwhitespace=true,
    frame=single,
    framerule=0.5pt,
    backgroundcolor=\color{gray!8},
    xleftmargin=12pt,
    xrightmargin=12pt,
    aboveskip=1.2em,
    belowskip=1.2em
]
SYSTEM:
You are a canonical user-turn interaction-intent classifier. Your task is to identify exactly one primary local interaction action performed by TARGET_USER_TURN in the given dialogue state.

Do not classify from keywords alone. First reconstruct the currently open task and the relation between the target user turn and the immediately preceding assistant turn. The raw target user turn is the final source of evidence. The profile, context, and payload view are auxiliary evidence only. Payload is not a primary intent label.

Apply the following rules in order and stop at the first matching rule.

A. SET_REGISTER
The user sets or invokes the assistant's role, identity, persona, protocol, language, tone, response register, or behavior mode for the current or future interaction. A purely local format, length, or style constraint is not SET_REGISTER by itself.

B. REPAIR
The user explicitly indicates that the previous assistant response, code, solution, interpretation, or behavior was wrong, failed, misunderstood the request, or violated a requirement. A bare traceback or error log is not REPAIR unless it is tied to the assistant's prior answer, code, or plan.

C. GROUND_ACCEPT
The user greets, thanks, accepts, confirms, closes, gives simple positive feedback, or asks to continue without adding a new task constraint. "Continue" or "Next" is GROUND_ACCEPT only when no new condition is added.

D. SUPPLY
The user mainly provides information, evidence, code, logs, text, numbers, preferences, choices, or an answer without requesting a new processing operation. "Here is the article" is SUPPLY, whereas "Here is the article; summarize it" introduces a task.

E. AMEND
The user advances or updates an already open task, output, or session procedure by adding or changing constraints, format, scope, style, length, focus, or the next item or step. AMEND does not require the user to claim that the assistant was wrong.

F. INITIATE
The user opens a new top-level goal, question, task, request, capability probe, or open discussion after all more specific labels are ruled out.

G. AMBIGUOUS
Use only when the primary action remains unstable after applying all rules. State the two most plausible competing labels in the rationale.

Decision order:
SET_REGISTER -> REPAIR -> GROUND_ACCEPT -> SUPPLY -> AMEND -> INITIATE -> AMBIGUOUS.

INPUT:
<USER_PROFILE>
{user_profile}
</USER_PROFILE>

<DIALOGUE_CONTEXT>
{dialogue_context}
</DIALOGUE_CONTEXT>

<LAST_ASSISTANT_TURN>
{last_assistant_turn}
</LAST_ASSISTANT_TURN>

<TARGET_USER_TURN>
{target_user_turn}
</TARGET_USER_TURN>

OUTPUT FORMAT:
{
  "label": "INITIATE",
  "step_applied": "F",
  "confidence": "high",
  "has_payload": "no",
  "payload_type": "none",
  "secondary_attribute": {
    "answer_space": "closed",
    "constraint_type": "NA",
    "repair_type": "NA",
    "acceptance_type": "NA",
    "register_scope": "NA"
  },
  "rationale": "The user opens a new explicit question."
}

Return exactly one JSON object and no additional text.
\end{lstlisting}

\FloatBarrier

\section{Intent-Calibrated Policy Optimization}
\label{app:grpo_details}

\subsection{Group-Relative Advantages}

For each directive-conditioned prompt $x_t$, \icr{} assigns a calibrated reward $r_i$ to
every sampled user turn $y_i$ in the group $\mathcal{Y}$. We optimize these rewards with a
group-relative update based on GRPO~\citep{shao2024deepseekmath}, whose clipped surrogate
is inherited from PPO~\citep{schulman2017ppo}. A frozen behavior policy
$\pi_{\theta_{\mathrm{old}}}$ samples $K$ candidates, and their rewards are standardized
within the group:

\begin{equation}
\bar{r} = \frac{1}{K}\sum_{j=1}^{K}r_j,
\qquad
s_r^2 = \frac{1}{K}\sum_{j=1}^{K}(r_j-\bar{r})^2,
\qquad
\widehat{A}_i = \frac{r_i-\bar{r}}{\sqrt{s_r^2+\varepsilon_{\mathrm{num}}}}.
\label{eq:grpo_group_advantage}
\end{equation}

Here $\varepsilon_{\mathrm{num}}>0$ ensures numerical stability. The group-relative
construction removes the need to train a separate value model. Rewards and normalized
advantages are treated as fixed targets during each policy update.


\subsection{Ordering Preservation under Group Normalization}

Whenever $s_r^2>0$, Equation~\ref{eq:grpo_group_advantage} applies the same strictly
increasing affine transformation to every reward in a group. Therefore, for any
intent-violating candidate $v$ and intent-compliant candidate $c$ in a mixed group,

\begin{equation}
r_v + m \leq r_c
\quad\Longrightarrow\quad
\widehat{A}_v < \widehat{A}_c.
\label{eq:grpo_order_preservation}
\end{equation}

Thus, group normalization cannot make a violating candidate outrank a compliant alternative. 
This statement concerns pairwise ordering.
Whether every violating candidate
has a negative normalized advantage also depends on the complete reward distribution in the
group and is therefore reported empirically. If all rewards are identical, every normalized
advantage is zero and the group contributes no directional policy-gradient signal.


\subsection{Clipped Policy Objective}

For token position $\ell$ in candidate $y_i$, define the importance ratio

\begin{equation}
\rho_{i,\ell}(\theta)
=
\frac{\pi_\theta(y_{i,\ell}\mid x_t,y_{i,<\ell})}
{\pi_{\theta_{\mathrm{old}}}(y_{i,\ell}\mid x_t,y_{i,<\ell})}.
\label{eq:grpo_importance_ratio}
\end{equation}

The clipped surrogate contribution is

\begin{equation}
\mathcal{L}^{\mathrm{clip}}_{i,\ell}(\theta)
=
\min\!\left(
\rho_{i,\ell}(\theta)\widehat{A}_i,
\operatorname{clip}\!\left(\rho_{i,\ell}(\theta),1-\varepsilon_{\mathrm{clip}},1+\varepsilon_{\mathrm{clip}}\right)\widehat{A}_i
\right).
\label{eq:grpo_clipped_surrogate}
\end{equation}

We maximize the response-length-normalized objective

\begin{equation}
\begin{aligned}
\mathcal{J}_{\mathrm{GRPO}}(\theta)
= \mathrm{E}_{x_t,\,\mathcal{Y}\sim\pi_{\theta_{\mathrm{old}}}}
\Bigg[
\frac{1}{K}\sum_{i=1}^{K}\frac{1}{|y_i|}
\sum_{\ell=1}^{|y_i|}
\Big(
\mathcal{L}^{\mathrm{clip}}_{i,\ell}(\theta)
-\beta d^{\mathrm{KL}}_{i,\ell}(\theta)
\Big)
\Bigg],
\end{aligned}
\label{eq:grpo_objective}
\end{equation}

where $d^{\mathrm{KL}}_{i,\ell}$ measures divergence from a frozen reference policy
$\pi_{\mathrm{ref}}$ and $\beta$ controls its strength. The same sequence-level advantage
is applied to all tokens in a candidate. Length normalization prevents long responses from
dominating the update, and clipping limits abrupt policy changes. \icr{} determines the
within-group ordering. The policy objective converts that ordering into a stable update.

\FloatBarrier

\section{Data Construction and Annotation Quality}
\label{app:data_details}

\subsection{Source Corpus and Turn Expansion}
We use the conversation-level train, validation, and test partitions of
LMSYS-USP~\citep{wang2025usp}. 
To formulate next-user simulation as a turn-level prediction task, we expand
each conversation into one example for every eligible user turn following \citet{naous2026flipping}.
For a target user turn $u_t$, the model input consists of the implicit profile
and the dialogue prefix ending with the immediately preceding assistant turn,
while $u_t$ serves as the generation target.
The conversation-level split is fixed before this expansion, ensuring that all
turns from the same conversation remain in the same partition.
After removing turns labeled \textsc{Ambiguous}, the resulting dataset contains
444,635 training examples, 27,778 validation examples, and 9,233 test examples.


\subsection{Corpus Size and Annotation Coverage}
To make the effective supervision scale explicit, we distinguish among raw
user turns, turns receiving a valid taxonomy label, and non-ambiguous turns
retained for six-way intent modeling. Table~\ref{tab:appendix_corpus_coverage}
reports these quantities for each data partition, allowing annotation coverage
and the final usable sample pool to be assessed separately.

\begin{table}[H]
\centering
\small
\setlength{\tabcolsep}{7pt}
\begin{tabular}{lrrrrrr}
\toprule
\textbf{Split}
& \textbf{Conversations}
& \textbf{User turns}
& \textbf{Labeled turns}
& \textbf{Missing labels}
& \textbf{Non-ambiguous}
& \textbf{Coverage (\%)} \\
\midrule
Train      & 87,882 & 446,106 & 446,087 & 19 & 444,635 & 99.996 \\
Validation & 4,626  & 27,871  & 27,869  & 2  & 27,778  & 99.993 \\
Test       & 2,366  & 9,260   & 9,259   & 1  & 9,233   & 99.989 \\
\midrule
\textbf{Total} & \textbf{94,874} & \textbf{483,237} & \textbf{483,215}
& \textbf{22} & \textbf{481,646} & \textbf{99.995} \\
\bottomrule
\end{tabular}
\caption{Corpus size and taxonomy coverage.
Coverage is the percentage of user turns for which the annotation pipeline produced a valid taxonomy label. Non-ambiguous counts exclude \textsc{Ambiguous} but retain all six canonical directives.
}
\label{tab:appendix_corpus_coverage}
\end{table}


\subsection{Intent Distribution}

Beyond aggregate annotation coverage, we examine how local interaction actions
are distributed across the training, validation, and test partitions.
Table~\ref{tab:appendix_intent_distribution} reports both the count and
within-split proportion of each directive, while retaining
\textsc{Ambiguous} only for annotation auditing. This breakdown makes the
split composition explicit and reveals class imbalance that may be obscured
by aggregate performance.

The distribution is intentionally not rebalanced: \textsc{Initiate} and \textsc{Amend}
reflect the dominant interaction patterns in the source corpus, while \textsc{Supply} is
comparatively rare. We therefore report both accuracy and macro-F1, and use
intent-stratified sampling for human verification and comparative preference evaluation.

\begin{table}[H]
\centering
\small
\setlength{\tabcolsep}{7pt}
\begin{tabular}{lrrrr}
\toprule
\textbf{Intent}
& \textbf{Train}
& \textbf{Validation}
& \textbf{Test}
& \textbf{Total} \\
\midrule
\textsc{Initiate}      & 174,959 (39.22\%) & 11,271 (40.44\%) & 3,758 (40.59\%) & 189,988 (39.32\%) \\
\textsc{Amend}         & 187,577 (42.05\%) & 11,850 (42.52\%) & 3,694 (39.90\%) & 203,121 (42.04\%) \\
\textsc{Supply}        & 8,642 (1.94\%)    & 414 (1.49\%)     & 225 (2.43\%)    & 9,281 (1.92\%) \\
\textsc{Repair}        & 22,837 (5.12\%)   & 1,473 (5.29\%)   & 431 (4.65\%)    & 24,741 (5.12\%) \\
\textsc{SetRegister}   & 27,894 (6.25\%)   & 1,469 (5.27\%)   & 737 (7.96\%)    & 30,100 (6.23\%) \\
\textsc{GroundAccept}  & 22,726 (5.09\%)   & 1,301 (4.67\%)   & 388 (4.19\%)    & 24,415 (5.05\%) \\
\textsc{Ambiguous}     & 1,452 (0.33\%)    & 91 (0.33\%)      & 26 (0.28\%)     & 1,569 (0.32\%) \\
\bottomrule
\end{tabular}
\caption{Intent distribution: count and percentage of labeled user turns.
Percentages use all successfully labeled turns in the corresponding split as the denominator.
}
\label{tab:appendix_intent_distribution}
\end{table}

\subsection{Human Validation of Intent Annotation and Prediction}
\label{app:human_intent_audit}

For the annotation-quality audit, we sample 1315 observed user turns from the
test set using intent-stratified sampling. Three human experts independently label
each anonymized turn using the canonical intent taxonomy without access to the
automatic labels. We evaluate both the retrospective verifier and the
prospective predictor against these expert reference labels.
The retrospective mode receives $(p,c_t,u_t)$ and measures whether an observed
turn can be assigned the intended canonical label.
The prospective mode receives only $(p,c_t)$ and measures agreement with the
intent selected in the observed human trajectory.

\begin{table}[t]
    \centering
    \small
    \setlength{\tabcolsep}{2pt}
    \begin{tabular}{lrrrrrr}
    \toprule
    \multirow{2}{*}{\textbf{Intent}}
    & \multicolumn{3}{c}{\textbf{Retrospective verifier}}
    & \multicolumn{3}{c}{\textbf{Prospective predictor}} \\
    \cmidrule(lr){2-4}
    \cmidrule(lr){5-7}
    & \textbf{P} & \textbf{R} & \textbf{F1}
    & \textbf{P} & \textbf{R} & \textbf{F1} \\
    \midrule
    
    \textsc{Initiate}
    & 88.28 & 88.95 & 0.886
    & 78.59 & 65.71 & 0.716 \\
    
    \textsc{Amend}
    & 88.78 & 86.31 & 0.875
    & 73.76 & 70.55 & 0.721 \\
    
    \textsc{Supply}
    & 70.37 & 89.06 & 0.786
    & 40.88 & 87.50 & 0.557 \\
    
    \textsc{Repair}
    & 91.67 & 83.33 & 0.873
    & 75.38 & 74.24 & 0.748 \\
    
    \textsc{SetRegister}
    & 90.83 & 90.83 & 0.908
    & 64.47 & 89.91 & 0.751 \\
    
    \textsc{GroundAccept}
    & 93.02 & 90.91 & \textbf{0.920}
    & 89.19 & 75.00 & \textbf{0.815} \\
    
    \midrule
    \textbf{Accuracy}
    & \multicolumn{3}{c}{\textbf{87.88}}
    & \multicolumn{3}{c}{71.39} \\
    
    \textbf{Macro-F1}
    & \multicolumn{3}{c}{\textbf{0.875}}
    & \multicolumn{3}{c}{0.718} \\
    
    \textbf{Model--human $\kappa$}
    & \multicolumn{3}{c}{\textbf{0.8223}}
    & \multicolumn{3}{c}{0.5950} \\
    \bottomrule
    \end{tabular}
    
    \caption{Per-intent agreement with expert labels. The retrospective verifier
    observes the realized target user turn, whereas the prospective predictor does
    not. Precision, recall, and accuracy are percentages, whereas F1 and Cohen's
    $\kappa$ are reported on a 0--1 scale. Cohen's $\kappa$~\citep{cohen1960coefficient}
    measures model--human agreement.}
    \label{tab:human_intent_detailed}
    \end{table}

\begin{table}[t]
    \centering
    \small
    \setlength{\tabcolsep}{7pt}
    \begin{tabular}{llr}
    \toprule
    \textbf{Expert label}
    & \textbf{Verifier prediction}
    & \textbf{Rate (\%)} \\
    \midrule
    \textsc{Initiate} & \textsc{Amend} & 7.81 \\
    \textsc{Amend} & \textsc{Initiate} & 10.32 \\
    \textsc{Repair} & \textsc{Amend} & 9.09 \\
    \bottomrule
    \end{tabular}
    \caption{
    \textbf{Most frequent retrospective-verifier confusions.}
    Rates are normalized by the corresponding expert source class.
    }
    \label{tab:human_intent_confusions}
\end{table}

The retrospective verifier's remaining errors concentrate on adjacent
taxonomy boundaries.
The most frequent confusions are
\textsc{Initiate}$\rightarrow$\textsc{Amend} (7.81\%),
\textsc{Amend}$\rightarrow$\textsc{Initiate} (10.32\%), and
\textsc{Repair}$\rightarrow$\textsc{Amend} (9.09\%).
These boundaries depend on whether the user opens a new top-level thread,
updates an existing task, or explicitly attributes the required change to an
assistant error.

The prospective predictor is necessarily more difficult because the target
user turn is unavailable and multiple next intents may be plausible.
Its high recall but lower precision on \textsc{Supply} and
\textsc{SetRegister} indicates that it sometimes proposes these intents in
states where the observed human trajectory selected another valid transition.
Accordingly, we interpret prospective accuracy as agreement with the observed
next intent, not as a claim that every disagreement is interactionally invalid.

\subsection{Human Intent Evaluation on Generated User Turns}
\label{app:human_generated_intent}
To test whether system-level improvements persist independently of the
automatic verifier, we evaluate generated user turns with expert intent labels.
We report an intent-balanced accuracy
$\mathrm{IA}_{\mathrm{bal}}$ and a reweighted accuracy
$\mathrm{IA}_{\mathrm{rw}}$.
The former gives each intent equal representation in the audit set, whereas
the latter reweights the human per-intent accuracies by the natural intent
distribution of the test set.

\begin{table}[t]
\centering
\small
\setlength{\tabcolsep}{7pt}
\begin{tabular}{lrrrr}
\toprule
\textbf{System}
& \textbf{Human}
  $\mathrm{IA}_{\mathrm{bal}}$
& \textbf{Human}
  $\mathrm{IA}_{\mathrm{rw}}$
& \textbf{Qwen}
  $\mathrm{IA}_{\mathrm{bal}}$
& \textbf{Qwen}
$\mathrm{IA}_{\mathrm{full}}$ \\
\midrule
USP w/ Directive
& 49.17 & 54.91 & 52.50 & 62.28 \\
\method{} w/o RL
& 84.17 & 79.96 & 87.50 & 81.98 \\
\method{}
& \textbf{90.83} & \textbf{87.14}
& \textbf{92.50} & \textbf{86.62} \\
\bottomrule
\end{tabular}
\caption{
\textbf{Intent accuracy on model-generated user turns under human and
automatic labels.}
}
\label{tab:human_generated_intent}
\end{table}

All four measurements produce the same system ordering:
$\method{}
>
\method{}\text{ w/o RL}
>
\mathrm{USP}$.
Under balanced human labels, \method{} reaches 90.83\% Intent Accuracy,
compared with 84.17\% for \method{} w/o RL and 49.17\% for USP.
After reweighting to the natural intent distribution, \method{} retains a
7.18-point advantage over its supervised variant and a 32.23-point advantage
over USP.
This result directly reduces the concern that \method{} merely optimizes for
idiosyncrasies of the training-time verifier.

\FloatBarrier

\section{Implementation and Evaluation Details}
\label{app:implementation}

\subsection{Training Configuration}

\begin{table}[H]
\centering
\small
\setlength{\tabcolsep}{6pt}
\begin{tabular}{p{0.20\textwidth} p{0.32\textwidth} p{0.32\textwidth}}
\toprule
\textbf{Configuration} & \textbf{Supervised Fine Tuning Phase} & \textbf{Reinforcement Learning Phase} \\
\midrule
Initialization
& LLaMA-3-8B Base
& Intent-SFT checkpoint \\

Training examples
& 444,635 canonical turns
& 49,844 canonical turns \\

Training duration
& 3 epochs
& 1 nominal epoch\\

Precision
& BF16
& BF16 \\

LoRA rank $r$
& 64
& 16 \\

LoRA $\alpha$
& 32
& 32 \\

LoRA dropout
& 0.05
& 0.05 \\

LoRA target modules
& \texttt{q,k,v,gate,up,}\newline
  \texttt{down\_proj,lm\_head}
& \texttt{q,k,v,gate,up,down\_proj} \\

Learning rate
& $5\times10^{-5}$
& $5\times10^{-7}$ \\

Effective batch
& 648 target turns
& 12 completions per update \\

Maximum sequence length
& 4,096 tokens
& 4,096 tokens \\

Group size $K$
& --
& 4 \\

KL coefficient $\beta$
& --
& 0.05 \\

Policy clip $\epsilon$
& --
& 0.2 \\

margin $m$
& --
& 0.10 \\

Reward scaling
& --
& Within-group normalization \\

Policy iterations per batch
& --
& 1 \\

Gradient accumulation
& 81
& 4 per trainer rank \\

Hardware
& 4$\times$ NVIDIA RTX 4090
& 4$\times$ NVIDIA RTX 4090 \\
\bottomrule
\end{tabular}
\caption{Training configuration.}
\label{tab:appendix_training_config}
\end{table}

\subsection{Rollout and Decoding Configuration}
\begin{table}[H]
    \centering
    \small
    \setlength{\tabcolsep}{7pt}
    \begin{tabular}{lrr}
    \toprule
    \textbf{Parameter} & \textbf{RL rollout} & \textbf{Final evaluation} \\
    \midrule
    Temperature & 0.8 & 0 (greedy) \\
    Top-$p$ & 0.95 & 1.0 \\
    Top-$k$ & 0 & Disabled \\
    Repetition penalty & 1.0 & 1.2 \\
    Maximum new tokens & 4,096 & 4,096 \\
    Random seed(s) & 42 & 42 \\
    \bottomrule
    \end{tabular}
    \caption{Rollout and evaluation decoding.}
    \label{tab:appendix_decoding}
\end{table}
\FloatBarrier

\section{Additional Evaluation Metrics and Results}
\label{app:additional_results}
\subsection{Metric Definitions}
\label{app:metric_definitions}


\paragraph{Turn-level intent realization.}
For $N$ evaluation instances with target interaction directives $z_i$ and
predictions $\hat{z}_i$ from the frozen canonical verifier, exact Intent
Accuracy is
\begin{equation}
\mathrm{Acc}_{\mathrm{intent}}
=
\frac{1}{N}
\sum_{i=1}^{N}
\mathbf{1}\!\left[\hat{z}_i=z_i\right].
\end{equation}

Let $\mathcal{Z}$ denote the six canonical directives:
\textsc{Initiate}, \textsc{Amend}, \textsc{Repair}, \textsc{Supply},
\textsc{GroundAccept}, and \textsc{SetRegister}. For directive
$k\in\mathcal{Z}$, let $P_k$ and $R_k$ denote its precision and recall.
Macro-F1, reported on a 0--1 scale, is the unweighted class average
\begin{equation}
\mathrm{MacroF1}
=
\frac{1}{|\mathcal{Z}|}
\sum_{k\in\mathcal{Z}}
\frac{2P_kR_k}{P_k+R_k},
\end{equation}
where a class with $P_k+R_k=0$ contributes zero. \textsc{Ambiguous} is
retained for audit accounting but is not treated as an additional class in
the macro average.

\paragraph{Reference-based semantic and stylistic fidelity.}
Let $\tilde{u}_i$ and $u_i$ be the generated and observed next-user turns,
respectively. Given frozen semantic and style encoders
$f_{\mathrm{sem}}$ and $f_{\mathrm{sty}}$, we compute
\begin{align}
\mathrm{SimCSE}
&=
\frac{1}{N}
\sum_{i=1}^{N}
\cos\!\left(
f_{\mathrm{sem}}(\tilde{u}_i),
f_{\mathrm{sem}}(u_i)
\right),\\
\mathrm{StyleCSE}
&=
\frac{1}{N}
\sum_{i=1}^{N}
\cos\!\left(
f_{\mathrm{sty}}(\tilde{u}_i),
f_{\mathrm{sty}}(u_i)
\right).
\end{align}
SimCSE measures semantic agreement with the observed user turn, whereas
StyleCSE measures similarity in user-side linguistic style. These metrics
are complementary: neither requires exact lexical overlap, and neither
directly verifies whether the requested interaction intent is realized.

\paragraph{Reference-free turn-level interaction quality.}
A frozen, model-blind LLM judge evaluates each candidate without access to
the observed user turn. The judge sees only the user profile, visible
dialogue state, immediately preceding assistant turn, target directive, and
anonymous candidate.

For instance $i$, let
$d_i$, $s_i$, $a_i$, $p_i$, and $u_i\in[0,5]$ denote the canonical scores
for directive realization, dialogue-state coherence, information
appropriateness, profile consistency, and user authenticity and economy,
respectively. After applying the pre-registered rubric consistency rules,
we define
\begin{align}
\mathrm{CtxValid}
&=
\frac{1}{N}
\sum_{i=1}^{N}
\frac{20}{3}\left(d_i+s_i+a_i\right),\\
\mathrm{UserAuth}
&=
\frac{1}{N}
\sum_{i=1}^{N}
\frac{20}{2}\left(p_i+u_i\right).
\end{align}

Contextual Validity therefore captures whether the candidate performs the
requested interaction intent coherently and with appropriate information.
User Authenticity captures profile consistency and natural, economical
user-side expression. Both aggregates are reported on a 0--100 scale.

\paragraph{Controlled-prefix trajectory metrics.}
For multi-turn evaluation, each checkpoint is generated under its original
gold dialogue prefix. Earlier generated turns are not recursively inserted
into later checkpoints. This controlled-prefix protocol isolates the
simulator's consistency across dialogue states from downstream error
propagation.

For trajectory $j\in\{1,\ldots,M\}$ containing $T_j$ evaluated checkpoints,
let
\begin{equation}
c_{jt}
=
\mathbf{1}\!\left[\hat{z}_{jt}=z_{jt}\right].
\end{equation}
Step Accuracy micro-averages correctness over all checkpoints:
\begin{equation}
\mathrm{Acc}_{\mathrm{step}}
=
\frac{\sum_{j=1}^{M}\sum_{t=1}^{T_j}c_{jt}}
     {\sum_{j=1}^{M}T_j}.
\end{equation}
Mean Trajectory Accuracy first computes the fraction of correct checkpoints
within each trajectory and then weights trajectories equally:
\begin{equation}
\mathrm{Acc}_{\mathrm{traj}}
=
\frac{1}{M}
\sum_{j=1}^{M}
\left(
\frac{1}{T_j}
\sum_{t=1}^{T_j}c_{jt}
\right).
\end{equation}
All-Turn Success is the fraction of trajectories for which every target
directive is realized correctly:
\begin{equation}
\mathrm{Success}_{\mathrm{all}}
=
\frac{1}{M}
\sum_{j=1}^{M}
\prod_{t=1}^{T_j}c_{jt}.
\end{equation}
The reported multi-turn SimCSE and StyleCSE scores are averaged over all
evaluated checkpoint turns.

\paragraph{Reference-free trajectory quality.}
For each controlled-prefix trajectory $j$, the session-level judge assigns
$r_j$, $h_j$, and $g_j\in[0,5]$ for role authenticity, interaction
performance, and goal progress, respectively. We report
\begin{align}
\mathrm{Role}
&=
\frac{20}{M}\sum_{j=1}^{M}r_j,\\
\mathrm{Interaction}
&=
\frac{20}{M}\sum_{j=1}^{M}h_j,\\
\mathrm{Goal}
&=
\frac{20}{M}\sum_{j=1}^{M}g_j,\\
\mathrm{Total}
&=
(\mathrm{Role}
+
\mathrm{Interaction}
+
\mathrm{Goal})/3.
\end{align}

Role measures whether a plausible and profile-consistent user identity is
maintained across checkpoints. Interaction measures local directive
realization, state tracking, and appropriate transitions between user
actions. Goal measures whether active goals and constraints are preserved
and advanced without unjustified abandonment or premature closure.

\subsection{Per-Directive Accuracy}
Table~\ref{tab:appendix_per_intent_accuracy} reports exact intent accuracy for each
target directive. Macro Avg. weights the six classes equally, while Weighted Acc. uses
their empirical frequencies in the evaluation set. Reporting both prevents common intents
from hiding rare-class failures while retaining performance on the natural test mixture.

\begin{table}[H]
    \centering
    \small
    \setlength{\tabcolsep}{5pt}
    \resizebox{\textwidth}{!}{%
    \begin{tabular}{lrrrrrrrr}
    \toprule
    \textbf{Model}
    & \textsc{Initiate}
    & \textsc{Amend}
    & \textsc{Supply}
    & \textsc{Repair}
    & \textsc{SetRegister}
    & \textsc{GroundAccept}
    & \textbf{Macro Avg.} 
    & \textbf{Weighted Acc.} \\
    \midrule
    LLaMA-3-8B Base + Directive & 65.9 & 42.3 & 18.8 & 65.2 & 80.7 & 79.5 & 58.7 & 56.2 \\
    USP + Directive & 76.0 & 60.5 & 32.8 & 42.4 & 39.4 & 50.0 & 50.2 & 62.3 \\
    UserLM + Directive & 49.3 & 54.0 & 18.8 & 21.2 & 14.7 & 38.6 & 32.8 & 45.0 \\
    Intent-SFT & \underline{89.7} & \underline{73.9} & \underline{70.3} & \underline{81.8} & \underline{85.3} & \underline{93.2} & \underline{82.4} & \underline{82.0} \\
    \method{} & \textbf{92.6} & \textbf{80.8} & \textbf{73.4} & \textbf{83.3} & \textbf{91.7} & \textbf{95.5} & \textbf{86.2}  & \textbf{86.7} \\
    \bottomrule
    \end{tabular}}
    \caption{Exact directive accuracy (\%) by target intent.}
    \label{tab:appendix_per_intent_accuracy}
\end{table}

\FloatBarrier

\subsection{Counterfactual Directive Intervention}
\label{app:counterfactual_intervention}

We construct a within-context intervention suite to measure whether a user
simulator can express different interaction directives without changing the
underlying dialogue state. Held-out states are frozen before model
generation and balanced across six domains, four dialogue-depth buckets, and
two levels of profile salience. A separate affordance screen retains only states
in which every canonical directive is contextually possible. Each evaluated
system then generates one next-user turn for every directive in every state,
yielding $48\times6=288$ generations per system. 

\paragraph{Directive realization.}
Let $\mathcal{C}$ be the 48 contexts and $\mathcal{Z}$ the six canonical
directives. For system $m$, let $\hat z_{c,z}^{(m)}$ be the frozen verifier's
prediction for the turn generated in context $c$ under requested directive
$z$. Directive Accuracy is
\begin{equation}
\mathrm{Acc}_{\mathrm{dir}}^{(m)}
=
\frac{1}{|\mathcal{C}||\mathcal{Z}|}
\sum_{c\in\mathcal{C}}
\sum_{z\in\mathcal{Z}}
\mathbf{1}\!\left[\hat z_{c,z}^{(m)}=z\right].
\end{equation}
Macro-F1 is the unweighted mean of the six directive-wise F1 scores, with
\textsc{Ambiguous} retained only as a possible verifier output and not as a
seventh target class.

\paragraph{Context-level controllability.}
For threshold $k\in\{4,5,6\}$, context-level success is
\begin{equation}
S_{\geq k}^{(m)}
=
\frac{1}{|\mathcal{C}|}
\sum_{c\in\mathcal{C}}
\mathbf{1}\!\left[
\sum_{z\in\mathcal{Z}}
\mathbf{1}\!\left[\hat z_{c,z}^{(m)}=z\right]
\geq k
\right].
\end{equation}
All-Six is $S_{\geq6}$. We additionally report realized-action coverage,
\begin{equation}
\mathrm{Coverage}^{(m)}
=
\frac{1}{|\mathcal{C}|}
\sum_{c\in\mathcal{C}}
\left|\left\{\hat z_{c,z}^{(m)}:z\in\mathcal{Z}\right\}\right|,
\end{equation}
whose maximum is six. Coverage distinguishes a simulator that produces several
different but imperfect actions from one that collapses all directives into the
same behavior.

\begin{table*}[t]
    \centering
    \scriptsize
    \setlength{\tabcolsep}{4pt}
    \renewcommand{\arraystretch}{1.08}
    \caption{Within-context intervention comparison.
    The profile describes a detail-oriented creative
    writer, and the previous assistant has just proposed a three-act outline for
    \emph{The Elementalists}. All systems receive the identical visible dialogue
    state. Within each target block, rows show USP w/ Dir., \method{} w/o RL, and \method{}. Only the requested next-user
    directive changes; long \textsc{Supply} and \textsc{SetRegister} outputs
    are excerpted for layout.}
    \label{tab:counterfactual_case_three_models}
    \begin{tabular}{p{0.475\textwidth}p{0.475\textwidth}}
    \toprule

    \textbf{Target: INITIATE}\par\vspace{2pt}
    \begin{tabular}{@{}p{0.12\textwidth}@{\hspace{4pt}}p{0.325\textwidth}@{}}
    \textbf{USP} &
    \emph{``Please generate a sample query and corresponding response pair for
    our dataset. Make sure they are relevant to the above description.''}
    \\[2pt]
    \textbf{Intent-SFT} &
    \emph{``write a story about a man who travels back in time to meet his
    ancestors''}
    \\[2pt]
    \textbf{\method{}} &
    \emph{``write a story about a man who travels back in time to meet his
    ancestors''}
    \end{tabular}
    &
    \textbf{Target: AMEND}\par\vspace{2pt}
    \begin{tabular}{@{}p{0.12\textwidth}@{\hspace{4pt}}p{0.325\textwidth}@{}}
    \textbf{USP} &
    \emph{``Please expand on Act I''}
    \\[2pt]
    \textbf{Intent-SFT} &
    \emph{``write it out fully''}
    \\[2pt]
    \textbf{\method{}} &
    \emph{``write it again but make sure there is no mention of earthquakes''}
    \end{tabular}
    \\

    \addlinespace[4pt]
    \midrule

    \textbf{Target: REPAIR}\par\vspace{2pt}
    \begin{tabular}{@{}p{0.12\textwidth}@{\hspace{4pt}}p{0.325\textwidth}@{}}
    \textbf{USP} &
    \emph{``Please expand on Act I''}
    \\[2pt]
    \textbf{Intent-SFT} &
    \emph{``this isn't what i asked for at all''}
    \\[2pt]
    \textbf{\method{}} &
    \emph{``this doesn't really fit my description, please try again''}
    \end{tabular}
    &
    \textbf{Target: GROUND\_ACCEPT}\par\vspace{2pt}
    \begin{tabular}{@{}p{0.12\textwidth}@{\hspace{4pt}}p{0.325\textwidth}@{}}
    \textbf{USP} &
    \emph{``ok''}
    \\[2pt]
    \textbf{Intent-SFT} &
    \emph{``Continue''}
    \\[2pt]
    \textbf{\method{}} &
    \emph{``continue''}
    \end{tabular}
    \\

    \addlinespace[4pt]
    \midrule

    \textbf{Target: SUPPLY}\par\vspace{2pt}
    \begin{tabular}{@{}p{0.12\textwidth}@{\hspace{4pt}}p{0.325\textwidth}@{}}
    \textbf{USP} &
    \emph{``Please expand on Act I''}
    \\[2pt]
    \textbf{Intent-SFT} &
    \emph{``write a story about a man named NAME\_8 who travels back in time
    to meet his ancestors''}
    \\[2pt]
    \textbf{\method{}} &
    \emph{``The Quantum Physics Based Magic System: Quantum mechanics plays a
    crucial role in shaping our universe at its most fundamental level. It
    describes the behaviour of matter and energy at subatomic scales,
    revealing phenomena such as superposition, entanglement, and
    non-locality~\ldots''}
    \end{tabular}
    &
    \textbf{Target: SET\_REGISTER}\par\vspace{2pt}
    \begin{tabular}{@{}p{0.12\textwidth}@{\hspace{4pt}}p{0.325\textwidth}@{}}
    \textbf{USP} &
    \emph{``Let us begin our query with a few definitions. First, we define
    `time' as the arrow of causality that flows from cause to effect. Second,
    we define `history' as the totality of events that ever happened in
    sequence~\ldots''}
    \\[2pt]
    \textbf{Intent-SFT} &
    \emph{``you will now write a story about a man named NAME\_8 who finds
    himself trapped inside his own body. He tries desperately to escape but
    fails every time. One day he finally succeeds and escapes his
    prison~\ldots''}
    \\[2pt]
    \textbf{\method{}} &
    \emph{``you will now write stories about a character named NAME\_8 who
    travels back in time to meet his ancestors. He communicates with them via
    letters he leaves behind at various points along his timeline. Write a
    story about him meeting his great-great-grandfather during World War
    One.''}
    \end{tabular}
    \\

    \bottomrule
    \end{tabular}
\end{table*}

\paragraph{Reference-free turn quality.}
An anonymized turn-level rubric scores directive realization
$s_{\mathrm{dir}}$, dialogue-state coherence $s_{\mathrm{state}}$, information
appropriateness $s_{\mathrm{info}}$, profile consistency $s_{\mathrm{prof}}$,
and user authenticity/economy $s_{\mathrm{auth}}$ on a 0--5 scale. We report
\begin{align}
    \mathrm{CtxValid}
    &=\frac{20}{3}
    \left(s_{\mathrm{dir}}+s_{\mathrm{state}}+s_{\mathrm{info}}\right),\\
    \mathrm{UserAuth}
    &=10\left(s_{\mathrm{prof}}+s_{\mathrm{auth}}\right).
\end{align}

\paragraph{Qualitative intervention matrix.}
Table~\ref{tab:counterfactual_case_three_models} shows all six \method{} interventions for
one shared state across three models.

\subsection{Recursive Closed-Loop Pilot}
\label{app:closed_loop_pilot}

We complement it with a diagnostic closed-loop pilot in which each generated user
turn is sent to the same Qwen3.5-9B assistant and the assistant response
is appended before the next user turn. A model-independent controller observes
the current dialogue and a private scenario state, then selects one of the six
canonical directives. It requests \textsc{Supply} when information is needed,
\textsc{Repair} after a grounded assistant error, \textsc{Amend} when the active
task should be refined, and \textsc{GroundAccept} only after the success
criteria are met. The simulator receives the selected directive and generates
the next user turn. Sessions contain at most six user turns.


We freeze 24 scenarios before user-model generation, balanced across six
domains and three difficulty levels. We evaluate LLaMA-3-8B Base w/ Dir, 
USP w/ Dir, and \method{}, producing 72 recursively generated sessions and
365 user turns in total. Unlike the fixed-prefix setting, trajectories diverge
after the first generated turn. 
This divergence is the behavior of interest rather than a
violation of the evaluation contract.

\paragraph{Metrics.}
Turn-level Directive Accuracy and Macro-F1 compare the controller directive
with the frozen verifier prediction for every generated user turn. A blind
session judge scores role authenticity $r_s$, interaction performance $i_s$,
and goal progress $g_s$ on a 0--5 scale. Let $d_s$ and $f_s$ indicate goal
drift and controller failure, respectively. For session $s$, we define
\begin{align}
\mathrm{GoalSuccess}_s
&=\mathbf{1}\!\left[g_s\geq4 \land d_s=0 \land f_s=0\right],\\
\mathrm{ConstraintSat}_s
&=\mathbf{1}\!\left[i_s\geq4 \land d_s=0\right],\\
\mathrm{SessionScore}_s
&=\frac{20}{3}\left(r_s+i_s+g_s\right).
\end{align}

\begin{table*}[t]
    \centering
    \small
    \setlength{\tabcolsep}{5pt}
    \resizebox{\textwidth}{!}{%
    \begin{tabular}{lccccccccc}
    \toprule
    \textbf{Simulator} &
    \textbf{Dir. Acc.}$\uparrow$ &
    \textbf{Macro-F1}$\uparrow$ &
    \textbf{Goal Succ.}$\uparrow$ &
    \textbf{Constraint}$\uparrow$ &
    \textbf{Premature}$\downarrow$ &
    \textbf{Role}$\uparrow$ &
    \textbf{Interaction}$\uparrow$ &
    \textbf{Goal}$\uparrow$ &
    \textbf{Total}$\uparrow$ \\
    \midrule
    LLaMA-3-8B Base w/ Dir.
        & 31.03 & 0.240 & 37.50 & 54.17 & 8.33 & 57.50 & 60.00 & 57.50 & 58.33 \\
    USP w/ Dir.
        & 19.35 & 0.129 & 25.00 & 37.50 & 12.50 & 49.17 & 50.83 & 47.50 & 49.17 \\
    \method{}
        & \textbf{51.20} & \textbf{0.411} & \textbf{54.17}
        & \textbf{58.33} & 12.50 & \textbf{64.17} & \textbf{64.17}
        & \textbf{62.50} & \textbf{63.61} \\
    \bottomrule
    \end{tabular}}
    \caption{Diagnostic recursive closed-loop evaluation.}
    \label{tab:closed_loop_pilot}
\end{table*}

\method{} improves over USP by +31.85 percentage points in directive accuracy
(95\% CI [14.17, 40.14]), +29.17 points in goal success
([4.17, 54.17]), and +14.44 points in session score
([4.72, 24.72]).

\subsection{Cross-Judge Groupwise Evaluation}
\label{app:groupwise_preference}

The turn-level rubric in the main paper evaluates each candidate independently.
We additionally conduct a reference-free groupwise comparison in which five
anonymous candidates under the same profile, dialogue state, and target intent
are presented jointly.
The candidates are generated by LLaMA-3-8B Base, UserLM, USP,
\method{} w/o RL, and \method{}, all with directives.
The observed user turn and model identities are hidden.

Each judge assigns every candidate an overall score from 0 to 5 while
considering intent realization, dialogue-state coherence, profile consistency,
naturalness, appropriate detail, and absence of assistant or control leakage.
The judge also returns explicit rankings.
The plotted values are mean per-candidate overall scores, not scores derived
from the rankings.

\begin{figure}[t]
    \centering
    \includegraphics[width=0.62\textwidth]
    {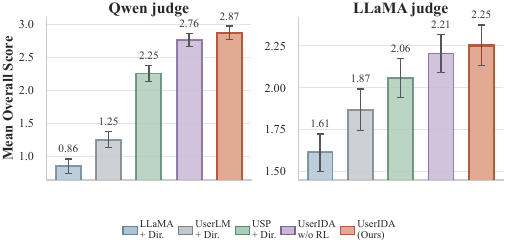}
    \caption{Cross-judge groupwise evaluation.}
    \label{fig:appendix_preference_scores}
\end{figure}

Both judges produce the same aggregate ordering:
\[
\mathrm{LLaMA}
<
\mathrm{UserLM}
<
\mathrm{USP}
<
\method{}\text{ w/o RL}
<
\method{}
\]
The Qwen-based judge assigns mean scores of
$0.86$, $1.25$, $2.25$, $2.76$, and $2.87$, respectively.
The LLaMA-based judge assigns
$1.61$, $1.87$, $2.06$, $2.21$, and $2.25$.
Although the judges use different absolute score calibrations, their system
ordering is identical.
We therefore interpret cross-judge rank consistency rather than directly
comparing score magnitudes across judges.
This analysis complements the human intent audit and the pointwise
reference-free metrics in the main paper.

\section{Prompt Templates and Model Interfaces}
\label{app:prompts}

\subsection{Intent-SFT and Controlled Inference Prompt}

Intent-SFT preserves the original user-simulator system prompt, profile, and dialogue
history. A turn-local system message is inserted immediately before the next user turn.
During training, the final user block contains the supervised target,
during controlled inference, the same block is left open for generation. 
The directive specifies the local interaction action rather than the wording of the target utterance.

\begin{lstlisting}[
    caption={Intent-SFT and controlled-inference prompt template.},
    label={lst:intent_sft_prompt},
    basicstyle=\ttfamily\small,
    breaklines=true,
    breakindent=0pt,
    breakatwhitespace=true,
    frame=single,
    framerule=0.5pt,
    backgroundcolor=\color{gray!8},
    xleftmargin=12pt,
    xrightmargin=12pt,
    aboveskip=1.2em,
    belowskip=1.2em
]
SYSTEM:
{original_user_simulator_system_prompt}

<USER_PROFILE>
{user_profile}
</USER_PROFILE>

The next user turn may be controlled by a user-intent signal. This signal is metadata for simulation and must not be copied into the user message.

Intent labels:
- INITIATE: The next user turn should open a new top-level goal, question, request, task, capability probe, or open discussion.
- AMEND: The next user turn should continue, update, or advance an existing open task, procedure, or dialogue thread without explicitly saying the assistant was wrong.
- REPAIR: The next user turn should explicitly indicate that the previous assistant answer, code, or plan was wrong, failed, misunderstood, or violated the user's requirement.
- SUPPLY: The next user turn should mainly provide information, evidence, code, logs, values, preferences, or payload without explicitly asking for a new processing task.
- GROUND_ACCEPT: The next user turn should be a short greeting, thanks, acceptance, confirmation, closing, or unconstrained continuation signal.
- SET_REGISTER: The next user turn should set or invoke a role, persona, protocol, register, or behavior mode for the assistant, either locally or persistently.

USER:
{earlier_user_turn_1}

ASSISTANT:
{earlier_assistant_turn_1}
...

USER:
{immediately_previous_user_turn}

ASSISTANT:
{immediately_previous_assistant_turn}

SYSTEM:
### User Intent Control for the NEXT user turn
Intent: {intent_label}
Guidance: {intent_definition}
Do not copy this control text into the user message.

USER: 
\end{lstlisting}

\subsection{Reference-Free Turn-Level Rubric Prompt}
\label{app:turn_level_judge_prompt}

The turn-level evaluator scores an anonymous candidate as a possible next
user turn under the visible dialogue state. It does not observe the gold
user turn or model identity, preventing reference overlap and model-specific
priors from directly determining the judgment.

The rubric separately assesses directive realization, dialogue-state
coherence, information appropriateness, profile consistency, and natural,
economical user-side expression. Explicit consistency rules cap scores for
wrong directives, assistant-role behavior, control leakage, truncation, and
empty output. Length alone is never rewarded.

\begin{lstlisting}[
    caption={Reference-free turn-level rubric prompt.},
    label={lst:turn_level_judge_prompt},
    basicstyle=\ttfamily\small,
    breaklines=true,
    breakindent=0pt,
    breakatwhitespace=true,
    frame=single,
    framerule=0.5pt,
    backgroundcolor=\color{gray!8},
    xleftmargin=12pt,
    xrightmargin=12pt,
    aboveskip=1.2em,
    belowskip=1.2em
]
SYSTEM:
You are a strict and meticulous evaluator of simulated NEXT USER turns.

Your task is to evaluate one generated message as a possible next USER turn
in the supplied dialogue state. The candidate is not an assistant response.
The model identity is hidden.

There may be multiple valid next-user messages. Do not assume a hidden
reference answer and do not reward phrase overlap with an imagined reference.
Judge only whether the candidate is interactionally valid, contextually
appropriate, profile-consistent, and natural for a human user.

==================================================
PROVIDED INFORMATION
==================================================

<PROFILE>
{profile}
</PROFILE>

<RECENT_DIALOGUE_CONTEXT>
{recent_context}
</RECENT_DIALOGUE_CONTEXT>

<PREVIOUS_ASSISTANT>
{previous_assistant}
</PREVIOUS_ASSISTANT>

<TARGET_DIRECTIVE>
Label: {target_intent}
Definition: {intent_definition}
Boundary check: {intent_specific_boundary_check}
</TARGET_DIRECTIVE>

<GENERATED_NEXT_USER_TURN>
{candidate}
</GENERATED_NEXT_USER_TURN>

==================================================
SCORING SCALE
==================================================

Score each dimension from 0 to 5. Half-point scores are allowed.

5.0 = Excellent: fully satisfies the dimension with no meaningful defect.
4.0 = Strong: correct with only a minor weakness.
3.0 = Acceptable: broadly valid but noticeably imperfect.
2.0 = Weak: major problem, although some relevant behavior remains.
1.0 = Severe failure: mostly invalid for the dimension.
0.0 = Completely invalid, empty, unintelligible, or wrong-role output.

Reserve scores above 4.0 for clearly strong outputs.

==================================================
DIMENSIONS
==================================================

1. directive_realization
Does the candidate realize the requested target interaction directive,
including its boundary relative to the other directives?

2. dialogue_state_coherence
Does the candidate correctly respond to the immediately previous assistant
turn, preserve the active task state and prior constraints, and avoid
contradiction, unsupported assumptions, or abrupt topic drift?

3. information_appropriateness
Does the candidate provide the information needed for this particular turn,
with appropriate specificity and without material omissions, unsupported
details, irrelevant elaboration, or unnecessary repetition?

4. profile_consistency
Is the candidate consistent with explicit profile facts, preferences,
behavioral tendencies, and speaking style?
Do not penalize a candidate merely because it does not explicitly mention
profile details when they are irrelevant.

5. user_authenticity_and_economy
Does the candidate sound like a natural human user's next message?
Penalize assistant-style service language, role reversal, synthetic templates,
control-label leakage, excessive formality, unnecessary structure, and
verbosity that is inappropriate for the current turn.
Length alone must never increase the score.

==================================================
INTENT-SENSITIVE NOTES
==================================================

- INITIATE should create a new top-level goal rather than continue the current open task.
- AMEND should update or extend an existing task without falsely claiming an assistant error.
- REPAIR should explicitly ground a correction in an assistant mistake, misunderstanding, failure, or violated requirement.
- SUPPLY should provide requested information or material. Long payloads are acceptable when the context requests them.
- GROUND_ACCEPT may be very short and should acknowledge, accept, continue, or close without adding a substantial new constraint.
- SET_REGISTER may naturally use imperative language to establish an ongoing role, tone, or protocol. Do not confuse valid protocol-setting language with control leakage.

==================================================
HARD RULES
==================================================

- If the candidate clearly performs a different directive, directive_realization must not exceed 1.5.
- If the candidate answers as the assistant, discusses hidden instructions, or emits system/control annotations, user_authenticity_and_economy must not exceed 1.0.
- If the candidate is empty or unintelligible, all scores must be 0.
- If the candidate is visibly truncated before completing its action, information_appropriateness and user_authenticity_and_economy must not exceed 1.5.
- Do not reward length, formatting, politeness, or profile keyword copying by themselves.

==================================================
MANDATORY CONSISTENCY CHECK BEFORE OUTPUT
==================================================

Apply these steps in order:

1. Classify the candidate's actually observed user action as predicted_directive.
2. Compare predicted_directive with the TARGET_DIRECTIVE label.
3. If they differ, set wrong_user_action=true and set directive_realization to
   1.5 or lower. This cap is mandatory even when the candidate is fluent,
   coherent, or performs its different action well.
4. Use 0.0 only when the candidate provides no usable evidence for that
   dimension (for example empty, unintelligible, or clearly wrong-role text).
   Otherwise use at least 0.5 and explain the defect.
5. Use 5.0 only for a rare, fully convincing result with no visible defect.
   If there is any minor weakness, use 4.5 or lower.
6. Verify that every score, flag, predicted_directive, and evidence sentence is
   mutually consistent before returning JSON.

==================================================
OUTPUT
==================================================

Return one valid JSON object and no other text. Do not use markdown or a code
fence. Every field shown below is required. predicted_directive must be exactly
one of: INITIATE, AMEND, REPAIR, SUPPLY, GROUND_ACCEPT, SET_REGISTER, AMBIGUOUS.

{
  "predicted_directive": "INITIATE",
  "scores": {
    "directive_realization": {
      "score": 0.0,
      "evidence": "One concise sentence."
    },
    "dialogue_state_coherence": {
      "score": 0.0,
      "evidence": "One concise sentence."
    },
    "information_appropriateness": {
      "score": 0.0,
      "evidence": "One concise sentence."
    },
    "profile_consistency": {
      "score": 0.0,
      "evidence": "One concise sentence."
    },
    "user_authenticity_and_economy": {
      "score": 0.0,
      "evidence": "One concise sentence."
    }
  },
  "flags": {
    "wrong_user_action": false,
    "assistant_role_behavior": false,
    "profile_contradiction": false,
    "unsupported_detail": false,
    "irrelevant_or_repetitive": false,
    "excessive_verbosity": false,
    "control_or_intent_leakage": false,
    "truncated_or_incomplete": false
  },
  "brief_summary": "One concise overall assessment."
}
\end{lstlisting}

\subsection{Controlled-Prefix Session-Level Rubric Prompt}
\label{app:session_level_judge_prompt}

The session-level evaluator scores role authenticity, interaction performance, and goal progress.
The evaluator first checks each candidate against its own visible state and
then assesses whether the simulator maintains a coherent user role,
interaction policy, and goal trajectory across checkpoints.

\begin{lstlisting}[
    caption={Controlled-prefix session-level rubric prompt.},
    label={lst:session_level_judge_prompt},
    basicstyle=\ttfamily\small,
    breaklines=true,
    breakindent=0pt,
    breakatwhitespace=true,
    frame=single,
    framerule=0.5pt,
    backgroundcolor=\color{gray!8},
    xleftmargin=12pt,
    xrightmargin=12pt,
    aboveskip=1.2em,
    belowskip=1.2em
]
SYSTEM:
You are a strict and meticulous evaluator of a simulated USER trajectory.

You will evaluate multiple generated next-user turns produced by one simulator
at different checkpoints from the same held-out conversation.

IMPORTANT CONTROLLED-PREFIX PROTOCOL:
Each checkpoint is evaluated under its own original gold dialogue prefix.
A generated user turn at one checkpoint was NOT fed into the assistant response
or into the next checkpoint. Therefore:

- Do not assume that a later assistant turn is responding to an earlier generated candidate.
- Do not penalize a candidate because a later gold assistant turn does not logically follow from it.
- Evaluate each generated user turn against its own visible dialogue state.
- Then evaluate whether the simulator shows a consistent user role, interaction policy, and goal trajectory across checkpoints.

The model identity is hidden. No gold user responses are provided.

==================================================
PROFILE
==================================================

<PROFILE>
{profile}
</PROFILE>

==================================================
CONTROLLED TRAJECTORY CHECKPOINTS
==================================================

{checkpoint_blocks}

==================================================
SCORING SCALE
==================================================

Score each session-level dimension from 0 to 5. Half-point scores are allowed.

5.0 = Excellent and consistently maintained across the trajectory.
4.0 = Strong with only minor localized weaknesses.
3.0 = Acceptable but inconsistent or imperfect at several checkpoints.
2.0 = Weak with major trajectory-level failures.
1.0 = Severe failure across most checkpoints.
0.0 = Unusable trajectory, mostly empty, unintelligible, or wrong-role output.

==================================================
SESSION-LEVEL DIMENSIONS
==================================================

1. role_authenticity
Evaluate whether the same plausible user is maintained across checkpoints.
Consider profile consistency, stable identity and speaking style, natural
variation, and absence of assistant-role behavior or mechanical profile
copying. Do not require explicit profile mentions when irrelevant.

2. interaction_performance
Evaluate correct interpretation of each previous assistant turn, local
directive realization and transitions, state tracking, efficient responses,
appropriate supply, and avoiding premature acceptance. Remember that
checkpoints use independent gold prefixes.

3. goal_progress
Evaluate preservation and advancement of goals and constraints, supplying
missing information, repairing mistakes, meaningful amendments, unresolved
concerns, and avoiding unjustified abandonment or premature closure.

==================================================
HARD RULES
==================================================

- If explicit profile facts are repeatedly contradicted, role_authenticity must not exceed 2.0.
- If most checkpoints realize the wrong local action, interaction_performance must not exceed 2.0.
- If the trajectory repeatedly abandons or prematurely closes unresolved goals, goal_progress must not exceed 2.0.
- Assistant-role output, control leakage, and repeated synthetic templates must reduce role_authenticity and interaction_performance.
- Length alone must never increase any score.
- A short acknowledgment can be optimal for GROUND_ACCEPT; a long requested payload can be appropriate for SUPPLY.

==================================================
OUTPUT
==================================================

Return one valid JSON object and no other text. Do not use markdown or a code
fence. Every field shown below is required. For each checkpoint,
predicted_directive must be one of INITIATE, AMEND, REPAIR, SUPPLY,
GROUND_ACCEPT, SET_REGISTER, AMBIGUOUS; local_validity must be one of valid,
partially_valid, invalid; main_issue must be one of none, wrong_action,
state_mismatch, profile_conflict, missing_information, premature_acceptance,
topic_drift, assistant_role, verbosity, other.

{
  "scores": {
    "role_authenticity": {
      "score": 0.0,
      "evidence": "One concise sentence summarizing trajectory-level evidence."
    },
    "interaction_performance": {
      "score": 0.0,
      "evidence": "One concise sentence summarizing trajectory-level evidence."
    },
    "goal_progress": {
      "score": 0.0,
      "evidence": "One concise sentence summarizing trajectory-level evidence."
    }
  },
  "checkpoint_diagnostics": [
    {
      "checkpoint_id": "1",
      "predicted_directive": "INITIATE",
      "local_validity": "valid",
      "main_issue": "none"
    }
  ],
  "trajectory_flags": {
    "persona_drift": false,
    "goal_drift": false,
    "repeated_wrong_action": false,
    "premature_closure": false,
    "unnecessary_repetition": false,
    "assistant_role_behavior": false,
    "control_leakage": false
  },
  "brief_summary": "One concise overall trajectory assessment."
}
\end{lstlisting}

\subsection{Comparative Groupwise Preference Prompt}
\label{app:preference_prompt}

The comparative evaluator ranks anonymous candidate next-user turns under the same
profile, dialogue state, and target directive. It jointly considers directive realization,
state coherence, profile consistency, naturalness, appropriate detail, and control
leakage. Length alone is not rewarded, and a fluent candidate with the wrong interaction
directive receives a low intent-fit score.

Each case is evaluated under two independently randomized candidate orders. Model
identities and the gold user turn are hidden.

\begin{lstlisting}[
    caption={Comparative groupwise preference prompt.},
    label={lst:groupwise_judge_prompt},
    basicstyle=\ttfamily\small,
    breaklines=true,
    breakindent=0pt,
    breakatwhitespace=true,
    frame=single,
    framerule=0.5pt,
    backgroundcolor=\color{gray!8},
    xleftmargin=12pt,
    xrightmargin=12pt,
    aboveskip=1.2em,
    belowskip=1.2em
]
SYSTEM:
You are a strict evaluator of simulated NEXT USER turns.
Your task is to compare multiple candidate next-user messages in the provided dialogue state. Model identities are hidden. Evaluate each candidate only as a possible next USER turn, not as an assistant response.

Target interaction intent: {target_intent}
Definition: {intent_definition}

Profile or system context:
{user_profile}
Recent dialogue context:
{dialogue_context}
Immediately previous assistant turn:
{last_assistant_turn}

Candidates:
Candidate A: {candidate_A}
Candidate B: {candidate_B}
Candidate C: {candidate_C}
Candidate D: {candidate_D}
Candidate E: {candidate_E}

Evaluate interaction-intent realization, dialogue-state coherence, profile consistency, natural user-like phrasing, user-likeness, appropriate detail, assistant or control leakage, and unsupported detail or templating. Do not reward length by itself. Do not infer model identity. A fluent candidate with the wrong interaction intent must score low on intent fit.

Return strict JSON only, without markdown:
{
  "candidates": [
    {
      "candidate_id": "A",
      "intent_fit": 0,
      "state_coherence": 0,
      "profile_consistency": 0,
      "naturalness": 0,
      "user_likeness": 0,
      "appropriate_detail": 0,
      "assistant_like_penalty": 0,
      "template_or_verbosity_penalty": 0,
      "overall_score": 0,
      "predicted_intent": "INITIATE",
      "brief_reason": "single-line reason"
    },
    ...
  ],
  "overall_ranking": ["A", "B", "C", "D", "E"],
  "intent_ranking": ["A", "B", "C", "D", "E"],
  "naturalness_ranking": ["A", "B", "C", "D", "E"],
  "best_candidate_id": "A",
  "all_candidates_bad": false,
  "confidence": "high"
}
All scores are numbers from 0 to 5. Return every candidate exactly once in the candidate records and rankings. Return no additional text.
\end{lstlisting}

\FloatBarrier


\end{document}